%% file: example_paper.tex
\documentclass[nohyperref]{article}

\usepackage{microtype}
\usepackage{graphicx}
\usepackage{caption}
\usepackage{subcaption}
\usepackage{booktabs} 

\usepackage{hyperref}

\usepackage[accepted]{icml2022}

\usepackage{bigints}
\usepackage{amsmath}
\usepackage{amssymb}
\usepackage{mathtools}
\usepackage{amsthm}

\usepackage[capitalize,noabbrev]{cleveref}

\input{math_commands.tex}

\theoremstyle{plain}
\newtheorem{theorem}{Theorem}[section]
\newtheorem{proposition}[theorem]{Proposition}

\newtheorem{corollary}[theorem]{Corollary}
\theoremstyle{definition}
\newtheorem{definition}[theorem]{Definition}

\theoremstyle{remark}

\RequirePackage{color} 
\RequirePackage{soul}
\setstcolor{blue}

\usepackage{ifthen}
\newboolean{draft}
\setboolean{draft}{true}
\newcommand{\authorcomment}[2]
{\ifthenelse
  {\boolean{draft}}
  {\hl{\{#1: #2\}}}
  {}
}

\usepackage[textsize=tiny]{todonotes}

\icmltitlerunning{Calibrated Learning to Defer with One-vs-All Classifiers}

\begin{document}

\twocolumn[
\icmltitle{Calibrated Learning to Defer with One-vs-All Classifiers}




\begin{icmlauthorlist}
\icmlauthor{Rajeev Verma}{uva}
\icmlauthor{Eric Nalisnick}{uva}
\end{icmlauthorlist}

\icmlaffiliation{uva}{Informatics Institute, University of Amsterdam, Amsterdam, Netherlands}

\icmlcorrespondingauthor{Rajeev Verma}{rajeev.ee15@gmail.com}
\icmlcorrespondingauthor{Eric Nalisnick}{e.t.nalisnick@uva.nl}

\icmlkeywords{Machine Learning, ICML}

\vskip 0.3in
]



\printAffiliationsAndNotice{}  

\begin{abstract}
The \textit{learning to defer} (L2D) framework has the potential to make AI systems safer.  For a given input, the system can defer the decision to a human if the human is more likely than the model to take the correct action. We study the calibration of L2D systems, investigating if the probabilities they output are sound.  We find that \citeauthor{pmlr-v119-mozannar20b}'s \citeyearpar{pmlr-v119-mozannar20b} multiclass framework is not calibrated with respect to expert correctness.  Moreover, it is not even guaranteed to produce valid probabilities due to its parameterization being degenerate for this purpose.  We propose an L2D system based on one-vs-all classifiers that is able to produce calibrated probabilities of expert correctness.  Furthermore, our loss function is also a consistent surrogate for multiclass L2D, like \citeauthor{pmlr-v119-mozannar20b}'s \citeyearpar{pmlr-v119-mozannar20b}.  Our experiments verify that not only is our system calibrated, but this benefit comes at no cost to accuracy.  Our model's accuracy is always comparable (and often superior) to \citeauthor{pmlr-v119-mozannar20b}'s \citeyearpar{pmlr-v119-mozannar20b} model's in tasks ranging from hate speech detection to galaxy classification to diagnosis of skin lesions.
\end{abstract}

\section{Introduction}
Machine learning is being deployed in ever more consequential and high-stakes tasks such as healthcare \citep{zoabi2021machine, kadampur2020skin}, criminal justice \citep{zhong2018legal, chalkidis-etal-2019-neural}, and autonomous driving \cite{https://doi.org/10.1002/rob.21918}.  Thus, the trust and safety of these systems is paramount \citep{DBLP:conf/iclr/HendrycksD19, nguyen2015deep}.  One near-term solution is to ensure a human is involved in the decision making process.  For example, \textit{learning with a rejection option} \cite{Chow1957AnOC} allows the model to abstain from making a decision, instead passing the burden to a human.  The decision to abstain or not is usually derived from the model's confidence.  For a self-driving car, a winding stretch of road could make the system unconfident in its abilities.  The system would then refuse to drive and forces the human to take control.  When the system becomes confident again (e.g.~on a straight road), it can then take back control from the human.  

\textit{Learning to defer} (L2D) \cite{10.5555/3327345.3327513} is another framework that supports machine-human collaboration.  In L2D, the human's confidence is modeled as well as the machine's.  This allows the system to compare the human's and model's expected performances.  Thus, L2D systems defer when \emph{the human is more likely than the model to take the correct action}.  Returning to the example of a self-driving car, an L2D system would pass control to the human only when it expects the human to drive better than itself.  In addition to safety, such behavior allows for an efficient \emph{division of labor} between the human and machine.  By knowing what the human knows, the model is free to adapt itself to complement the human.  The model can concentrate on performing easy tasks well if it knows a human can be relied upon for harder tasks.  

Most previous work has attempted to improve the overall accuracy of L2D systems.  However, if these systems are to be used in safety-critical scenarios, then other factors such as trust, transparency, and fairness are important as well \citep{10.5555/3327345.3327513}. \citet{Tschandl2020HumancomputerCF} found that AI systems can mislead physicians into incorrect diagnoses, even when the doctor is initially confident.  To help prevent such scenarios, we want our systems to be well \textit{calibrated}. The output probabilities should reflect the true uncertainties of the model and human.  In other words, the L2D system should be a good forecaster.  If the system says the expert has a $70\%$ chance of being correct, then the expert should indeed be correct in about $70$ out of $100$ cases. 


In this paper, we study the calibration of L2D systems.  We focus on \citeauthor{pmlr-v119-mozannar20b}'s \citeyearpar{pmlr-v119-mozannar20b} formulation since it is the only consistent surrogate loss for multiclass L2D.  We find that the \citet{pmlr-v119-mozannar20b} loss results in models that are not well-calibrated with respect to expert correctness.  The problem is intrinsic: the softmax parameterization allows the estimator to be \emph{greater than one}.  We propose an alternative loss based on one-vs-all classifiers that does not have this issue.  We use the method of \textit{error correcting output codes} \citep{Ramaswamy2018ConsistentAF} to show the multiclass L2D problem reduces to multiple binary classification problems. In turn, our one-vs-all surrogate is a consistent loss function, thus making it a superior alternative to \citeauthor{pmlr-v119-mozannar20b}'s \citeyearpar{pmlr-v119-mozannar20b} loss.  In experiments ranging from hate speech detection to galaxy classification to diagnosis of skin lesions, our model always performs comparably, if not better than, the \citet{pmlr-v119-mozannar20b} formulation in addition to other L2D frameworks (e.g. \citet{okati2021differentiable}) and common baselines (e.g. confidence thresholds).

\section{Background: Multiclass Learning To Defer}\label{sec:bg}
\citet{pmlr-v119-mozannar20b} proposed the only known consistent (surrogate) loss function for multiclass \textit{learning to defer} (L2D).  Hence, for much of this paper, we focus on their formulation.  We discuss other related work in Section \ref{sec:related}.  We provide a technical overview of L2D in this section before moving on to our innovations in subsequent sections.

\paragraph{Data} We first define the data for multiclass L2D.  Let $\mathcal{X}$ denote the feature space, and let $\mathcal{Y}$ denote the output space, which we will always assume to be a categorical encoding of multiple ($K$) classes.  We assume that we have samples from the true generative process: $\rvx_n \in \mathcal{X}$ denotes a feature vector, and $\ry_n \in \mathcal{Y}$ denotes the associated class defined by $\mathcal{Y}$ (1 of $K$).  The L2D problem also assumes that we have access to (human) expert demonstrations.  Denote the expert's prediction space as $\mathcal{M}$, which is usually taken to be equal to the label space: $\mathcal{M}$ = $\mathcal{Y}$.  The expert may also have access to additional information unavailable to the model.  The expert demonstrations are denoted $\rsm_n \in \mathcal{M}$ for the associated features $\rvx_n$.  The combined N-element training sample is $\mathcal{D} = \{\vx_n, y_n, m_n\}_{n=1}^{N}$.

\paragraph{Models} Turning to the models, \citeauthor{pmlr-v119-mozannar20b}'s \citeyearpar{pmlr-v119-mozannar20b} L2D framework is built from the classifier-rejector approach \citep{46544,NIPS2016_7634ea65}.  The goal is to learn two functions: the \textit{classifier}, $h: \mathcal{X} \rightarrow \mathcal{Y}$, and the \textit{rejector}, $r: \mathcal{X} \rightarrow \{0,1\}$.  When $r(\rvx)=0$, the classifier makes the decision in the typical way.  When $r(\rvx)=1$, the classifier abstains and defers the decision to a human (or other backup system).  The rejector can be interpreted as a meta-classifier, determining which inputs are appropriate to pass to $h(\rvx)$.


\paragraph{Learning} The learning problem requires fitting both the rejector and classifier.  When the classifier makes the prediction, then the system incurs a loss $\ell(h(\vx), y)$.  When the human makes the prediction (i.e.~$r(\vx)=1$), then the system incurs a loss $\ell_{\text{exp}}(m, y)$.  Using the rejector to combine these losses, we have the overall classifier-rejector loss: 
\begin{equation}\begin{split}
    L(&h, r) = \\ & \mathbb{E}_{\rvx, \ry, \rsm}\left[(1-r(\rvx)) \  \ell(h(\rvx), \ry) \ + \ r(\rvx) \ \ell_{\text{exp}}(\rsm, \ry) \right]
\end{split}
\end{equation} where the rejector is acting as an indicator function that controls which loss to use.  While this formulation is valid for general losses, the canonical $0-1$ loss is of special interest for classification tasks:
\begin{equation}\label{eq:0-1}\begin{split}
 &L_{0-1}(h, r) = \\ & \ \ \ \ \ \mathbb{E}_{\rvx, \ry, \rsm}\left[(1-r(\rvx)) \  \mathbb{I}[h(\rvx) \ne \ry] \ + \ r(\rvx) \ \mathbb{I}[\rsm \ne \ry] \right]
\end{split}
\end{equation} where $\mathbb{I}$ denotes an indicator function that checks if the prediction and label are equal or not.

\paragraph{Softmax Surrogate} The key innovation of \citet{pmlr-v119-mozannar20b} is the proposal of a consistent surrogate loss for $L_{0-1}$.  They accomplish this by first unifying the classifier and rejector via an augmented label space that includes the rejection option.  Formally, this label space is defined as $\mathcal{Y}^{\bot} = \mathcal{Y} \cup \{\bot\}$ where $\bot$ denotes the rejection option.  Secondly, \citet{pmlr-v119-mozannar20b} use a reduction to cost sensitive learning that ultimately resembles the cross-entropy loss for a softmax parameterization.  Let $g_{k}:\mathcal{X} \mapsto \mathbb{R}$ for $k \in [1, K]$ where $k$ denotes the class index, and let $g_{\bot}:\mathcal{X} \mapsto \mathbb{R}$ denote the rejection ($\bot$) option.  These $K+1$ functions are then combined in the following softmax-parameterized surrogate loss:
\begin{equation}\label{eq:sm_loss}\begin{split}
     \rphi_{\text{SM}}(g_{1}&,\ldots, g_{K}, g_{\bot}; \vx, y, m) =  \\ & -\log \left( \frac{\exp\{g_{y}(\vx)\}}{\sum_{y' \in \mathcal{Y}^{\bot}} \exp\{g_{y'}(\vx)\} }\right) \\ & - \mathbb{I}[m = y]  \ \log \left( \frac{\exp\{g_{\bot}(\vx)\}}{\sum_{y' \in \mathcal{Y}^{\bot}} \exp\{g_{y'}(\vx)\} }\right).
\end{split}
\end{equation} The intuition is that the first term maximizes the function $g_{k}$ associated with the true label.  The second term then maximizes the rejection function $g_{\bot}$ but only if the expert's prediction is correct.  At test time, the classifier is obtained by taking the maximum over $k \in [1, K]$: $\hat{y} = h(\vx) = \argmax_{k \in [1, K]} g_{k}(\vx)$.  The rejection function is similarly formulated as $r(\vx) = \mathbb{I}[ g_{\bot}(\vx) \ge \max_{k} g_{k}(\vx) ]$.  In practice, \citet{pmlr-v119-mozannar20b} introduce a hyperparameter $\alpha \in \mathbb{R}^{+}$ that re-weights the classifier loss when the expert is correct.  Using $\alpha < 1$ encourages a higher degree of division of labor between classifier and expert.  Yet for all $\alpha \ne 1$, the surrogate is no longer consistent.

The function $\rphi_{\text{SM}}$ is the first convex (in $g$) consistent surrogate loss proposed for L2D \citep{pmlr-v119-mozannar20b}.  The minimizers $g^{*}_{1},\ldots, g^{*}_{K}, g^{*}_{\bot}$ of $\rphi_{\text{SM}}$ also uniquely minimize $L_{0-1}(h, r) $, the $0-1$ loss from Equation \ref{eq:0-1}.  The resulting optimal classifier and rejector satisfy: 
\begin{equation}\begin{split}\label{eq:Bayes_optimal_rej_clf_0-1_loss}
    h^{*}(\vx) &= \argmax_{y \in \mathcal{Y}} \ \mathbb{P}(\ry = y | \vx), \\
    r^{*}(\vx) &= \mathbb{I}\left[\mathbb{P}(\rsm = \ry | \vx) \ge \max_{y \in \mathcal{Y}} \mathbb{P}(\ry = y | \vx) \right],
\end{split}
\end{equation} where $\mathbb{P}(\ry | \vx)$ is the probability of the label under the data generating process, and $\mathbb{P}(\rsm = \ry | \vx)$ is the probability that the expert is correct.  Recall that, by assumption, the expert likely will have additional knowledge not available to the classifier.  This assumption is what allows the expert to possibly outperform the Bayes optimal classifier.



\section{Problem with Softmax Parameterization}\label{sec:problem_with_softmax_parameterization}
The minimizers of the surrogate proposed by \citet{pmlr-v119-mozannar20b} should correspond to the Bayes optimal classifier and rejector.  In this section, we investigate if the resulting model can correctly estimate the underlying probability that the expert is correct.  We find that, unfortunately, the resulting models are not well calibrated.  The problem lies in the softmax parameterization: it yields a degenerate estimate of $\mathbb{P}(\rsm = \ry | \rvx)$.  Specifically, the estimator is unbounded, taking on values larger than one.  We do not study the calibration of the classifier since post-hoc methods (e.g.~temperature scaling \citep{10.5555/3305381.3305518}) can be applied to the classifier sub-components of both our method and \citeauthor{pmlr-v119-mozannar20b}'s \citeyearpar{pmlr-v119-mozannar20b}.   

\begin{figure*}
     \vskip 0.2in
     \centering
     \begin{subfigure}[b]{0.3\textwidth}
         \centering
         \includegraphics[width=\linewidth,height=5.0cm]{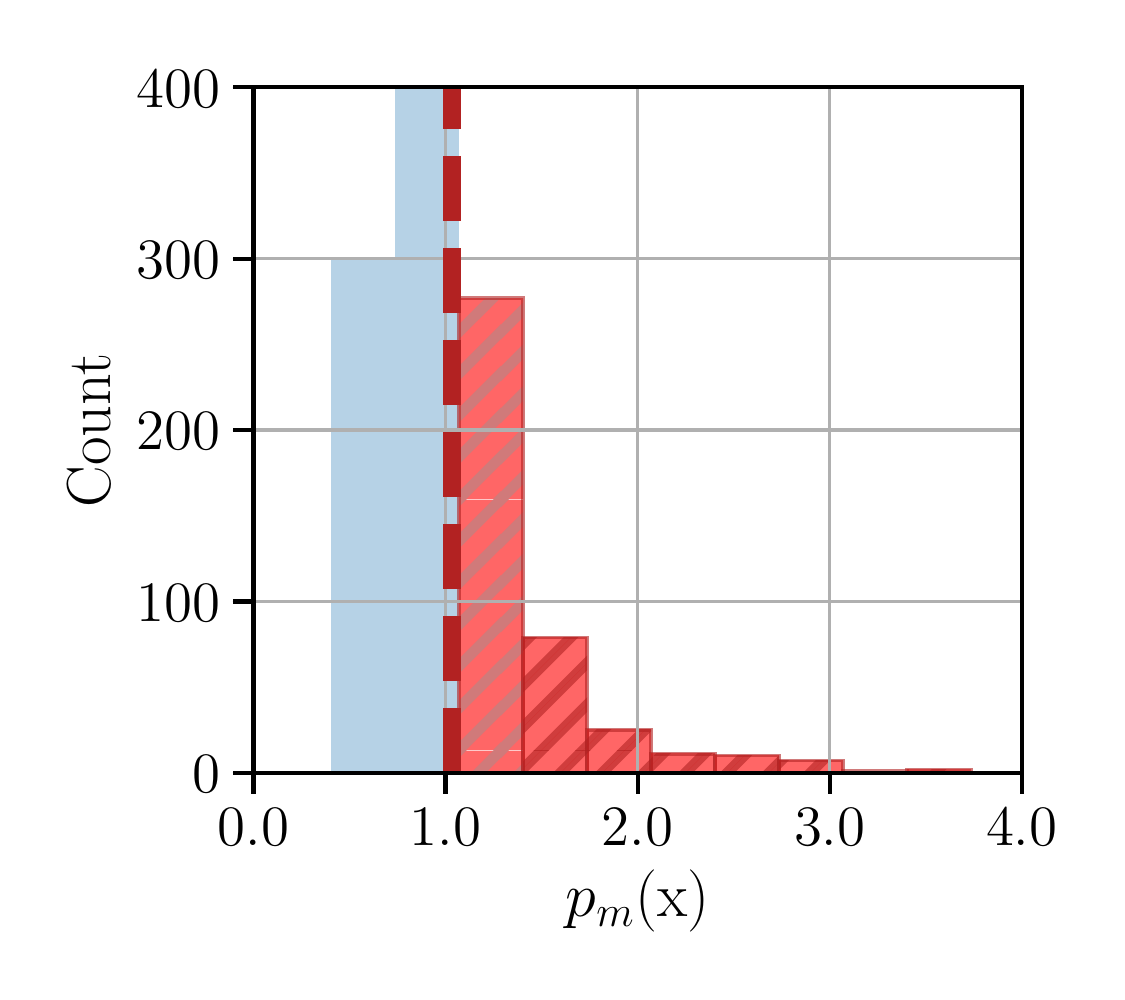}
         \caption{Empirical Distribution of $p_{\rsm}$ Values}
         \label{fig:sontag_confidence_pathology}
     \end{subfigure}
     \hfill
     \begin{subfigure}[b]{0.3\textwidth}
         \centering
         \includegraphics[width=\linewidth, height=5.575cm]{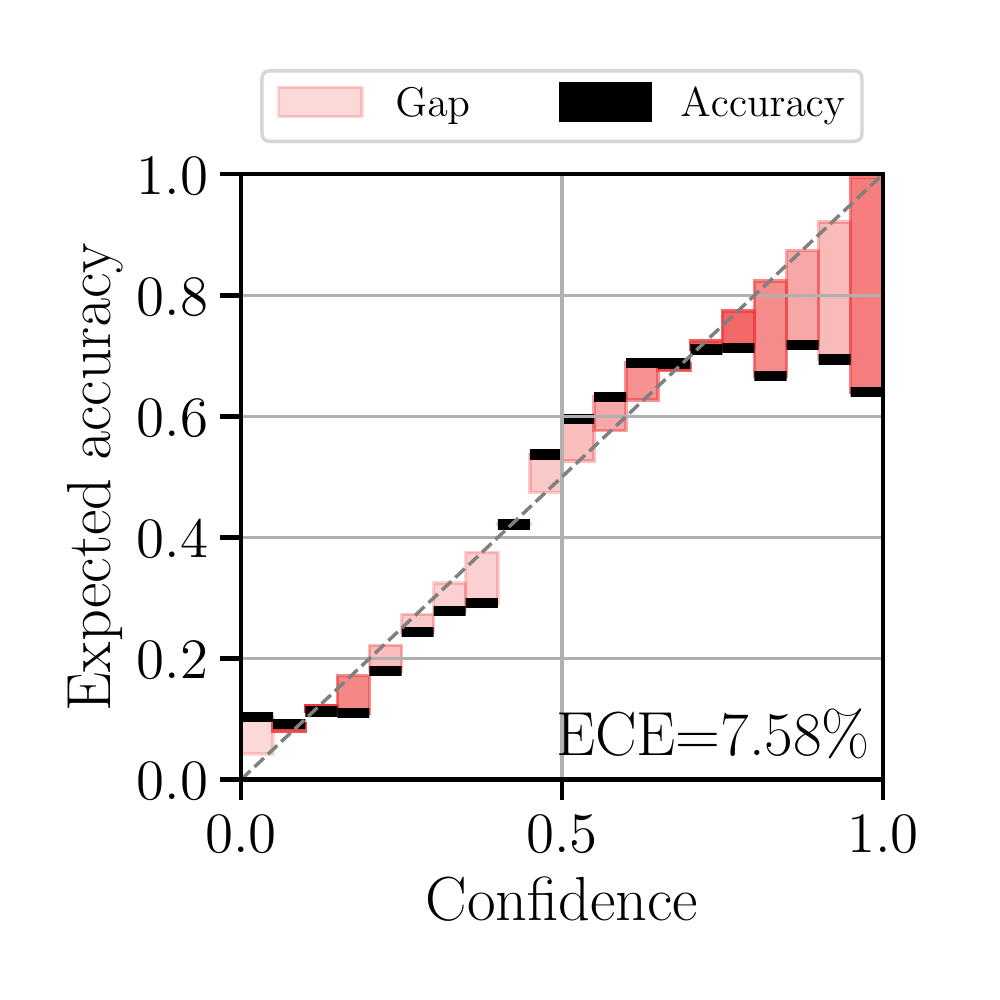}
         \caption{Reliability Diagram and ECE}
         \label{fig:sontag_rejector_reliability}
     \end{subfigure}
      \hfill
     \begin{subfigure}[b]{0.3\textwidth}
         \centering
         \includegraphics[width=\linewidth]{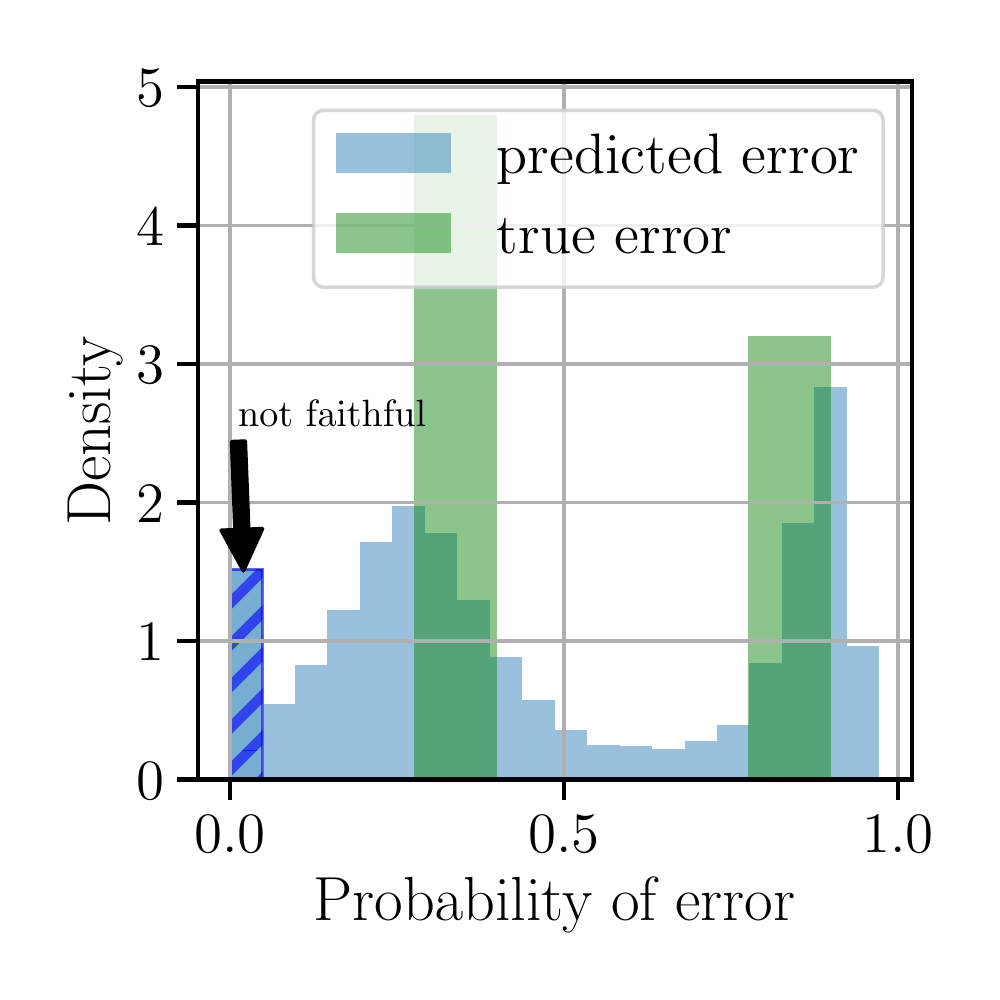}
         \caption{Empirical Distribution of $1 - p_{\rsm}(\rvx)$}
         \label{fig:sontag_true_error_vs_predicted_error}
     \end{subfigure}
     \hfill
     \caption{\textit{Calibration of Softmax Parameterization on CIFAR-10}: Subfigure (a) reports the observed values of $p_{\rsm}(\rvx)$ on the CIFAR-10 simulation study.  We find that $39.4\%$ of test samples have $p_{\rsm}(\rvx)>1$ (denoted in red).  Subfigure (b) reports a reliability diagram and the expected calibration error (ECE) when $p_{\rsm}(\rvx)$ is restricted to $(0, 1]$.  The shade of the bin color represents the proportion of samples in the bin (darker shade, more samples). Subfigure (c) shows the distribution of risk estimates.  Note the clear bias towards zero error. }
     \label{fig:sontag}
        \vskip -0.2in
\end{figure*}

\paragraph{Probabilistic Rejector} We first introduce the probabilistic rejection function.  One may be tempted to work directly with the deferral function from Equation \ref{eq:sm_loss}:
\begin{equation}\label{eq:def_func} p_{\bot}(\vx) =   \frac{\exp\{g_{\bot}(\vx)\}}{\sum_{y' \in \mathcal{Y}^{\bot}} \exp\{g_{y'}(\vx)\}}.
\end{equation} However, inspecting \citeauthor{pmlr-v119-mozannar20b}'s \citeyearpar{pmlr-v119-mozannar20b} Theorem 1, we see that $p^{*}_{\bot}(\vx) = \mathbb{P}(\rsm = \ry | \rvx)/(1+\mathbb{P}(\rsm = \ry | \rvx))$ at the Bayes optimum.  Rearranging this equation gives the appropriate estimator for $\mathbb{P}(\rsm = \ry | \rvx)$: 
\begin{equation}\label{eq:expert_prob}\begin{split}
    p_{\rsm}(\vx) & =    \frac{p_{\bot}(\vx)}{1- p_{\bot}(\vx)}.
\end{split}
\end{equation} The full derivation is in Appendix \ref{sec:probs_from_softmax_outputs}. A crucial observation is that $p_{\rsm}(\vx) \in (0, \infty)$, meaning that the function is unbounded from above.  This will be of consequence when considering if it is calibrated.

\paragraph{Calibration} We next define the relevant notion of calibration.  For the function $p_{\rsm}(\vx)$ from Equation \ref{eq:expert_prob}, we call $p_{\rsm}$ \textit{calibrated} if, for any confidence level $c \in (0, 1)$, the actual proportion of times the expert is correct is equal to $c$: \begin{equation}\begin{split}
     \mathbb{P}(\rsm = \ry \ | \  p_{\rsm}&(\vx) = c) = c.
\end{split}
\end{equation} This statement should hold for all possible instances $\vx$ with confidence $c$.  Since expert correctness is a binary classification problem, distribution calibration, confidence calibration, and classwise calibration all coincide \cite{pmlr-v89-vaicenavicius19a}.  



\paragraph{Calibration of Expert Correctness} We next examine if Equation \ref{eq:expert_prob} is a valid estimator of the probability that the expert's prediction is correct.  Unfortunately, $p_{\rsm}(\vx)$ is unbounded; we formalize this fact in the statement below.
\begin{proposition}\label{lemma:uncalibrated}
If $\exists \ \vx \in \mathcal{X}$ for which $p_{\bot}(\vx) > 1/2$, then $p_{\rsm}(\vx) > 1$. Hence $p_{\rsm}(\vx)$ cannot estimate $\mathbb{P}(\rsm = \ry | \vx)$.\end{proposition}
This proposition is obvious from the fact that $p_{\rsm}(\vx)$ is the odds of $p_{\bot}(\vx) \in (0,1)$. Proposition \ref{lemma:uncalibrated} does \textit{not} imply a problem with the consistency of \citeauthor{pmlr-v119-mozannar20b}'s \citeyearpar{pmlr-v119-mozannar20b} surrogate loss. Rather, it means that the softmax parameterization admits many solutions that do not correspond to valid estimators for $\mathbb{P}(\ry = \rsm|\vx)$.  In other words, the Bayes solutions seem to be `fragile' in the sense that they require $p_{\bot}(\vx) \le 1/2$ while its true range is $(0, 1)$.  


To make matters concrete, consider the case in which the expert is always correct, $\mathbb{P}(\rsm = \ry | \vx) =1$, while the class distribution is maximally entropic, $\mathbb{P}(\ry | \vx) = 1/K$.  From Equation \ref{eq:expert_prob}, a perfect expert implies that $p_{\bot}(\rvx) = 1/2$.  In turn, $p_{k}(\rvx) = 1 / (2K)$.  For $K = 2$, the softmax in Equation \ref{eq:sm_loss} would produce the vector $[1/4, \ 1/4, \ 1/2]$.  While the resulting model would indeed correctly defer to the expert (since $g_{\bot} > g_{k}$), the output is not what we might expect for a case in which the classifier is useless and the expert is an oracle.  Intuition suggests that we should see an output like $[\epsilon/2, \ \epsilon/2, \ 1 - \epsilon]$ where $\epsilon$ is a small positive constant, as this seems to more accurately reflect the expert's clear superiority.  In practice, perhaps optimization is finding well-performing but non-optimal solutions like this one.


\paragraph{Experimental Confirmation}  We now establish that $p_{\bot}(\vx) > 1/2$ does occur in practice.  We use a CIFAR-10 simulation that is similar to \citeauthor{pmlr-v119-mozannar20b}'s \citeyearpar{pmlr-v119-mozannar20b} CIFAR-10 experiment.  The expert is assumed to have non-uniform expertise: $75\%$ chance of being correct on the first five classes, and $20\%$ (i.e.~random) chance on the last five classes. Subfigure \ref{fig:sontag_confidence_pathology} shows a histogram of the values of $p_{\rsm}(\vx)$ as observed on the CIFAR-10 test set.  The blue bars represent the values less than or equal to one. The red bars show the pathological cases greater than one.  $39.4 \%$ of the test samples ($3940$ instances) resulted in $p_{\rsm}(\vx) > 1$.  

We also consider modifying $p_{\rsm}(\vx)$ so that all values greater than one are rounded down to one.  In this case, since now $p_{\rsm}(\vx)$ is forcibly restricted to $(0, 1]$, we can perform standard evaluations of calibration, such as plotting a reliability diagram and computing \textit{expected calibration error} (ECE).  In this case, the relevant ECE is defined as $$\text{ECE}(p_{\rsm}) = \mathbb{E}_{\rvx}|\mathbb{P}\left( \rsm = \ry \  | \ p_{\rsm}(\rvx) = c \right) - c|.$$  Subfigure \ref{fig:sontag_rejector_reliability} shows the reliability diagram and reports the ECE for confidence calibration when $p_{\rsm}(\vx)$ is restricted.  Unsurprisingly, we still observe that the model's estimate of the expert's correctness is uncalibrated, exhibiting overconfidence.  The ECE is $7.58\%$.  For comparison, our one-vs-all method has an ECE of $3.01\%$, as we will describe later.  Subfigure \ref{fig:sontag_true_error_vs_predicted_error} plots the distribution of risks: $1 - p_{\rsm}(\rvx)$.  Due to the probabilities being clamped to one, we see a false mode at zero error.  In turn, the system is not transparent about the actual risk that decision makers would encounter.

\paragraph{Proxy via Deferral Function}  Returning to Equation \ref{eq:def_func}, it is possible that the deferral function $p_{\bot}(\rvx)$ is a useful estimator of $\mathbb{P}(\rsm = \ry | \vx)$, despite that theory suggests otherwise.  Here the range is no longer a problem because $p_{\bot}(\rvx) \in (0, 1)$.  Moreover, as discussed in the example above, intuition suggests that $p_{\bot}(\rvx)$ should correlate with the expert's degree of superiority to  the classifier.  In the experiments (Section \ref{sec:cifar10_exp}), we investigate if the proxy $p_{\bot}$ is a useful estimator of $\mathbb{P}(\rsm = \ry | \vx)$.  We ultimately find that it is not, as it results in ECEs above $30\%$.

\paragraph{Applicability of Post-Hoc Techniques} There are a range of post-hoc techniques designed to fix mis-calibration in classifiers, e.g.~\textit{temperature scaling} \citep{10.5555/3305381.3305518}, \textit{Dirichlet calibration} \citep{kull2019temperature}, \textit{top-label calibration} \citep{Gupta2021ToplabelC}. These techniques employ a calibration map \citep{pmlr-v89-vaicenavicius19a}: a usually simple transformation that is applied to the confidence estimates to re-calibrate them. Such a map is fitted on a held-out validation set using some goodness-of-fit measure, e.g.~log-likelihood. Due to the softmax's range problem and interdependence of its $g$ functions, we do not know of a general procedure for defining and fitting a calibration map for the L2D setting.

\begin{figure*}
     \centering
       \begin{subfigure}[b]{0.275\textwidth}
         \centering
         \includegraphics[width=\linewidth, height=5.0cm]{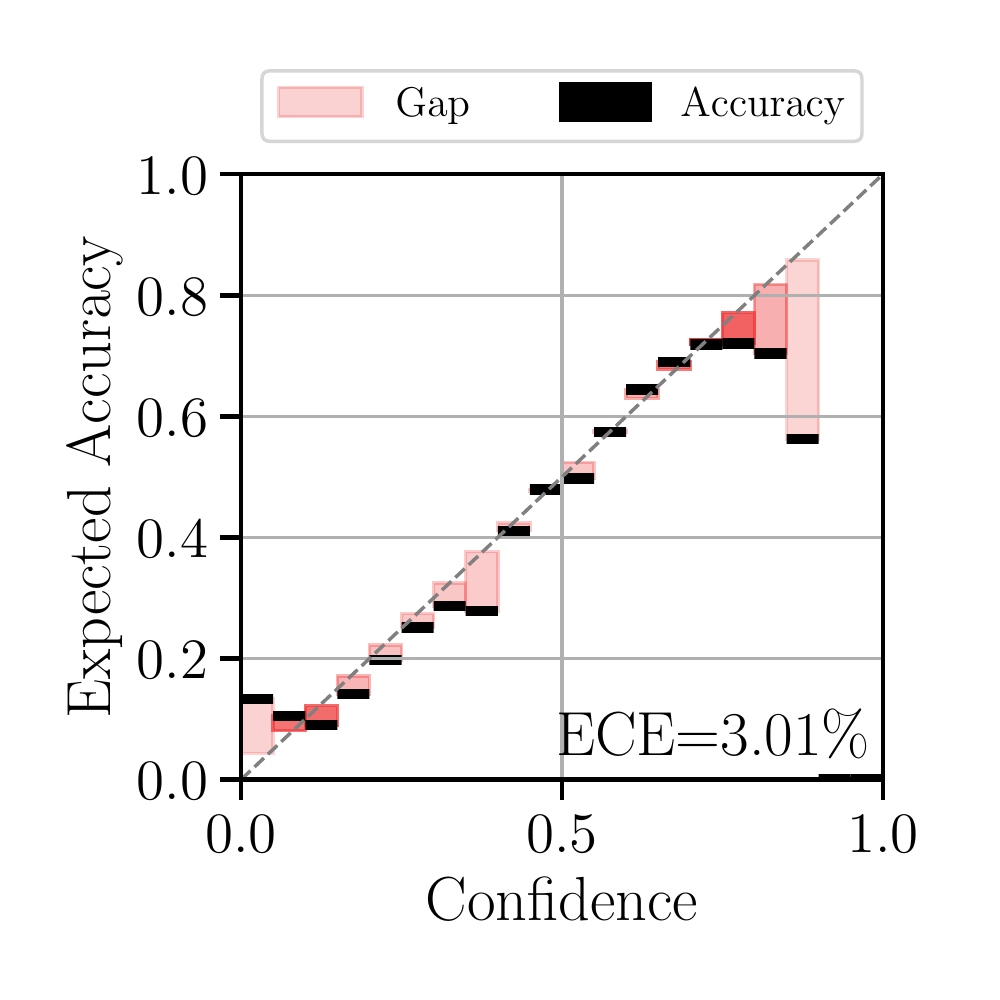}
         \caption{Reliability Diagram and ECE}
         \label{fig:OvA_rejector_reliability}
     \end{subfigure}
     \hfill
     \begin{subfigure}[b]{0.275\textwidth}
         \centering
         \includegraphics[width=\linewidth, height=5.0cm]{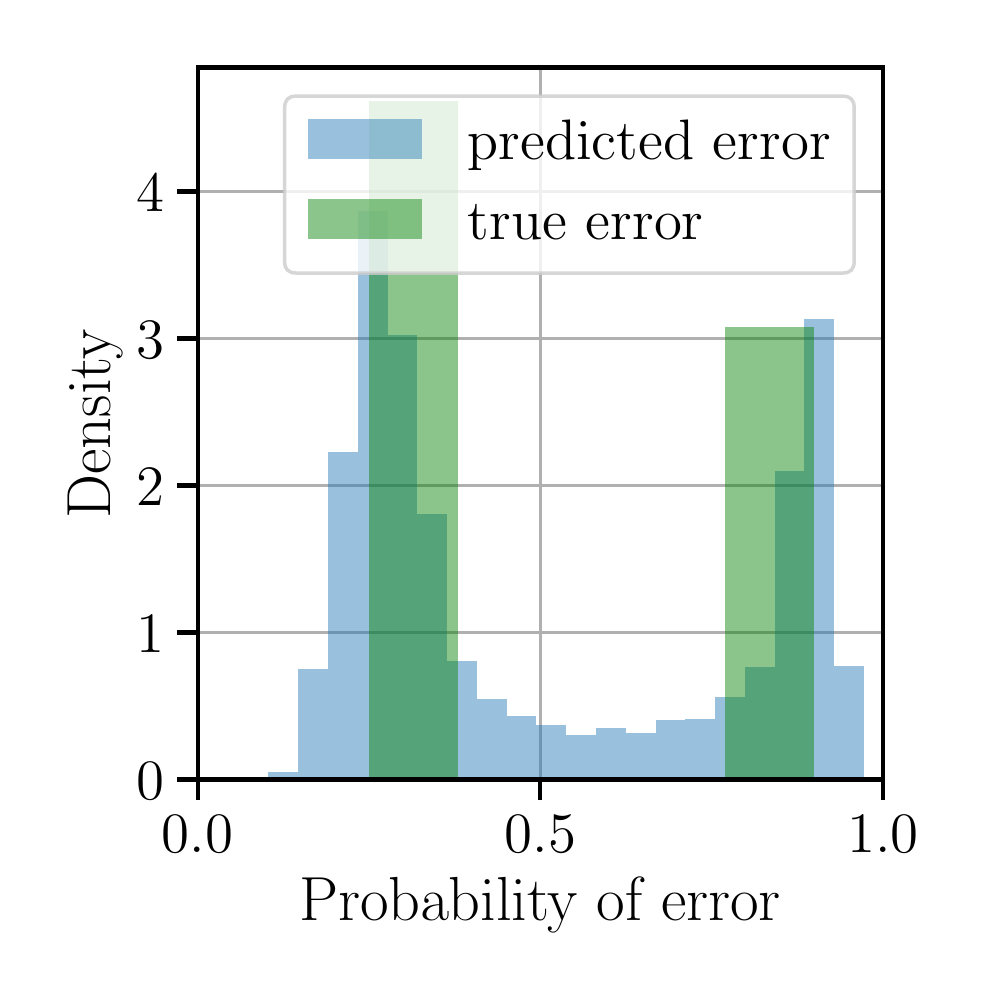}
         \caption{Empirical Dist.~of $1 - p^{\text{\tiny{OvA}}}_{\rsm}(\vx)$}
         \label{fig:OvA_true_error_vs_predicted_error}
     \end{subfigure}
     \hfill
     \begin{subfigure}[b]{0.435\textwidth}
         \centering
         \includegraphics[width=\linewidth, height=5.0cm]{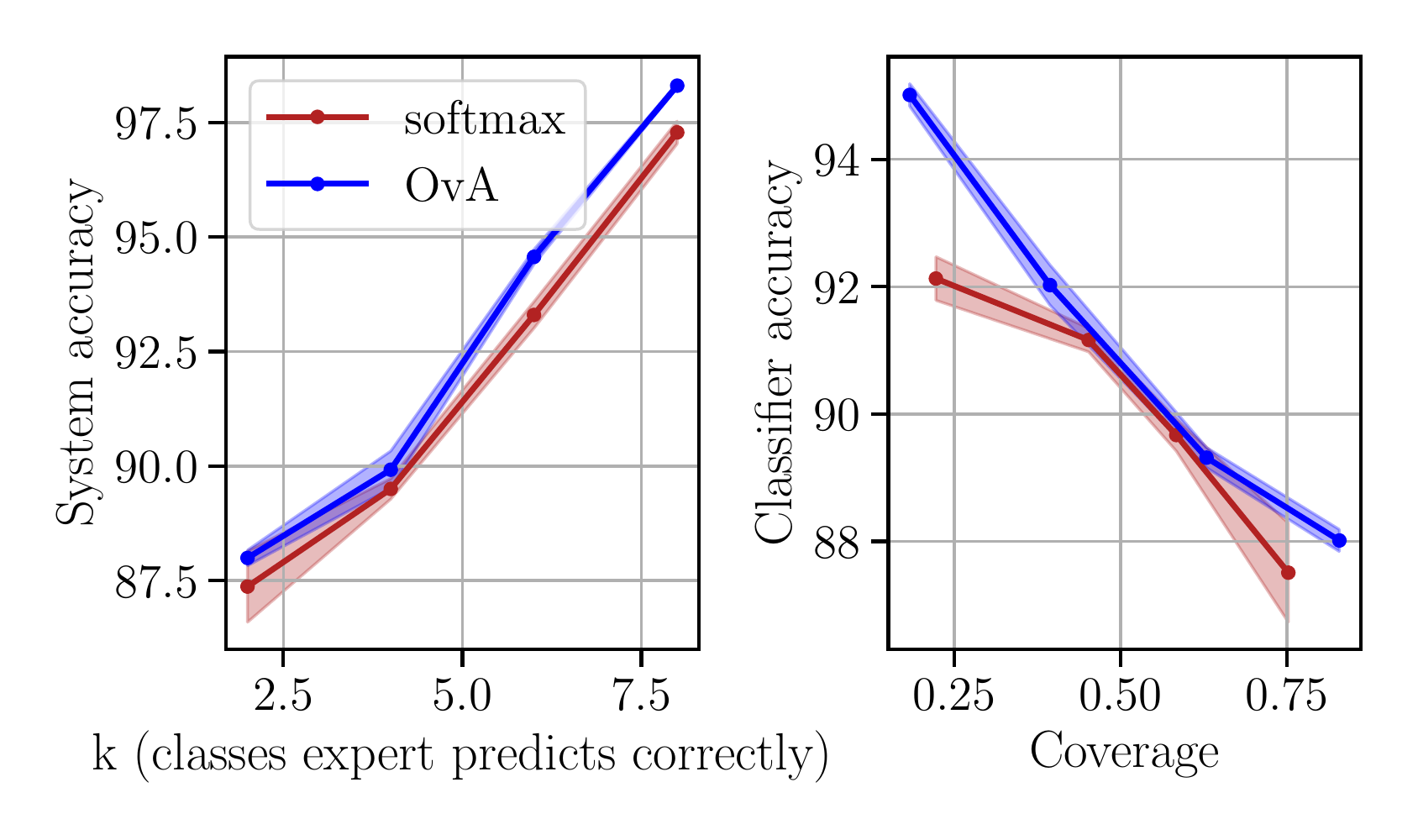}
         \caption{Accuracy and Coverage}
         \label{fig:coverage_accuracy_tradeoff}
     \end{subfigure}
     \label{fig:cifar-10}
     \caption{\textit{Calibration and Accuracy of OvA Parameterization on CIFAR-10}: Subfigure (a) reports a reliability diagram and the expected calibration error (ECE) for $p^{\text{\tiny{OvA}}}_{\rsm}(\vx)$ (Eq.~\ref{eq:expert_prob_ova}).  Darker bin shade means more samples in the bin. Subfigure (b) shows the distribution of risk estimates.  Subfigure (c) reports the accuracy as a function of an expert with increasing expertise (left) and of varying coverage (right). }
\end{figure*}

\section{Consistent and Calibrated L2D with a One-vs-All Surrogate Loss}\label{sec:OvA_surrogate_loss_for_L2D}
Given the difficulties in calibrating the softmax parameterization, we now consider an alternative.   We propose a one-vs-all parameterization (a.k.a.~one-vs-rest).  We show that the accompanying loss function is calibrated as well as a consistent surrogate for the $0-1$ loss.  Thus, our novel loss enjoys the same benefits as \citeauthor{pmlr-v119-mozannar20b}'s \citeyearpar{pmlr-v119-mozannar20b} formulation without its drawbacks.  

\subsection{One-vs-All-Based Surrogate Loss}
We propose the following one-vs-all-based surrogate for the same L2D problem described in Section \ref{sec:bg}.  Again assume we have $K+1$ functions $g_{1}(\rvx),\ldots, g_{K}(\rvx), g_{\bot}(\rvx)$ such that $g: \mathcal{X} \mapsto \mathbb{R}$.  And again, we observe training data of the form $\mathcal{D} = \{\vx_n, y_n, m_n\}_{n=1}^{N}$.  Our one-vs-all (OvA) surrogate loss takes the following point-wise form:
\begin{equation}\label{eq:ova_loss}\begin{split}
     & \rpsi_{\text{OvA}}(g_{1},\ldots, g_{K}, g_{\bot}; \vx, y, m) =  \\ & \ \ \  \rphi[g_{y}(\vx)] + \sum_{y' \in \mathcal{Y}, y' \ne y} \rphi[-g_{y'}(\vx)] \ \ + \\ & \ \ \   \rphi[-g_{\bot}(\vx)] + \mathbb{I}[m=y]\left(\rphi[g_{\bot}(\vx)] - \rphi[-g_{\bot}(\vx)]\right)
\end{split}
\end{equation} where $\rphi:\{\pm 1\}\times \mathbb{R} \mapsto \mathbb{R}_{+}$ is a binary surrogate loss.  For instance, when $\rphi$ is the logistic loss, we have $\rphi[f(\vx)] = \log(1 + \exp\{-f(\vx)\})$.  Our formulation is the OvA analog of \citeauthor{pmlr-v119-mozannar20b}'s \citeyearpar{pmlr-v119-mozannar20b} softmax-based loss.  The $g$-functions are entirely the same; the difference is in how they are combined.  Moreover, the classifier and rejector are computed exactly the same as in the softmax case: $ h(\vx) = \argmax_{k \in [1, K]} g_{k}(\vx)$, $r(\vx) = \mathbb{I}[ g_{\bot}(\vx) \ge \max_{k} g_{k}(\vx) ]$. In the experiments, we found no need for a re-weighting parameter that is analogous to $\alpha$ in \citeauthor{pmlr-v119-mozannar20b}'s \citeyearpar{pmlr-v119-mozannar20b} loss. One can be introduced similarly by re-weighting the first two terms in Equation \ref{eq:ova_loss} when the expert is correct.  \citet{10.5555/1005332.1005336} found OvA classifiers to work just as well as other approaches for multiclass classification, but in terms of general performance, OvA might falter in cases of data scarcity or severe class imbalance.  

We next turn to the probabilistic formulation of the rejector and classifier.  Starting with the former, the OvA formulation directly estimates the probability that the expert is correct: \begin{equation}\label{eq:expert_prob_ova}
    \mathbb{P}(\rsm = \ry | \vx) \ \approx \ p^{\text{\tiny{OvA}}}_{\rsm}(\vx) \ = \ (1 + \exp\{-g_{\bot}(\vx)\})^{-1}.
\end{equation}  $p^{\text{\tiny{OvA}}}_{\rsm}$ has the appropriate range of $(0, 1)$.  Moving on to the classifier, the foremost downside of the OvA formulation is that we can no longer compute normalized probabilities for all classes.  Rather, we can estimate only the probability of the most likely class: \begin{equation}\begin{split}\label{eq:OvA_clf_prob}
    \max_{k \in [1, K]} \mathbb{P}(\ry = k & | \vx) \ \approx \ \max_{k \in [1, K]} \ p^{\text{\tiny{OvA}}}_{k}(\vx) \\ & = \  \max_{k \in [1, K]} \ (1 + \exp\{-g_{k}(\vx)\})^{-1}.
\end{split}
\end{equation} Hence, we can evaluate the confidence calibration of the OvA classifier but not its distribution calibration.  This is a worthwhile trade off for having an appropriate estimator for $\mathbb{P}(\rsm = \ry | \vx)$ since distribution calibration is nearly impossible to achieve anyway \citep{zhao2021calibrating}. Multiclass-to-binary reduction has been shown to be an effective calibration strategy for traditional classifiers \citep{Gupta2021ToplabelC}.

\subsection{Theoretical Analysis}
We now justify the OvA loss by showing that, like \citeauthor{pmlr-v119-mozannar20b}'s \citeyearpar{pmlr-v119-mozannar20b} loss, ours is a consistent surrogate for the $0-1$ L2D loss (Equation \ref{eq:0-1}).  On one hand, this result is not surprising since our loss is the natural OvA-analog of the softmax-based loss.  However, we cannot construct our consistency proof in the same direct manner as \citet{pmlr-v119-mozannar20b}.  When we differentiate with respect to a particular $g(\vx)$, the other $g$'s drop from the OvA loss (but not from the softmax loss).  We proceed instead by the method of \textit{error correcting output codes} (ECOC) \citep{10.5555/1622826.1622834, Langford05sensitiveerror, 10.1162/15324430152733133, pmlr-v35-ramaswamy14}, a general technique for reducing multiclass problems to multiple binary problems.  We sketch the approach here and provide the details in Appendix \ref{sec:OvA_surrogate_loss_is_calibrated}.  

ECOC requires that we construct a coding matrix, which for our case is $\mathbf{M} \in \{-1, +1\}^{K \times (K+1)}$ with $K$ being the number of classes in the multiclass problem.  Each column then corresponds to a binary problem.  The entries of the matrix are determined as follows.  The $K \times K$ sub-matrix $\mathbf{M}_{1:K, 1:K}$ has $+1$ along its diagonal and $-1$ on the off-diagonal.  The entries in the $K+1$-th column are given by the function $m_{y, K+1}(\rsm) = (-1 + 2 \mathbb{I}[y = \rsm])$.  Now that we have constructed the coding matrix, we use Equation 1 from \citet{Ramaswamy2018ConsistentAF} to derive the closed form expression of the surrogate loss in Appendix \ref{sec:OvA_closed_form_expression} (Equation \ref{eq:ova_loss}). 
We then arrive at our final result: \begin{theorem}\label{thm: OvA_loss_calibration}
For a strictly proper binary composite loss $\rphi$ with a well-defined continuous inverse link function $\rgamma^{-1}$, $\rpsi_{\text{OvA}}$ (Equation \ref{eq:ova_loss}) is a calibrated surrogate for the $0-1$ learning to defer loss (Equation \ref{eq:0-1}).
\end{theorem}
The complete proof is in Appendix \ref{sec:OvA_surrogate_loss_is_calibrated}.
We also provide background information on calibration and consistency in Appendix \ref{sec:learning_theory_preliminaries}, which includes a discussion of proper binary composite losses.   
Lastly, assuming \textit{minimizability} \citep{Steinwart2007HowTC}---i.e.~that our hypothesis class is sufficiently large (all measurable functions)---the calibration result from Theorem \ref{thm: OvA_loss_calibration} implies consistency. \begin{corollary}\label{cor:OvA_loss_consistent}
Assume that $g \in \mathcal{F}$, where $\mathcal{F}$ is the hypothesis class of all measurable functions.  \textit{Minimizability} \citep{Steinwart2007HowTC} is then satisfied for $\rpsi_{\text{OvA}}$, and it follows that $\rpsi_{\text{OvA}}$ is a consistent surrogate for the $0-1$ learning to defer loss (Equation \ref{eq:0-1}).
\end{corollary} 

 Thus, $\rpsi_{\text{OvA}}$ is also a consistent loss function for L2D. This means that the minimizer of the proposed loss function $\rpsi_{\text{OvA}}$ over all measurable functions agrees with the Bayes optimal classifier and rejector (Equation \ref{eq:Bayes_optimal_rej_clf_0-1_loss}).

\begin{figure*}
     \centering
       \begin{subfigure}{0.45\textwidth}
         \centering
         \includegraphics[width=\linewidth, height=6.75cm]{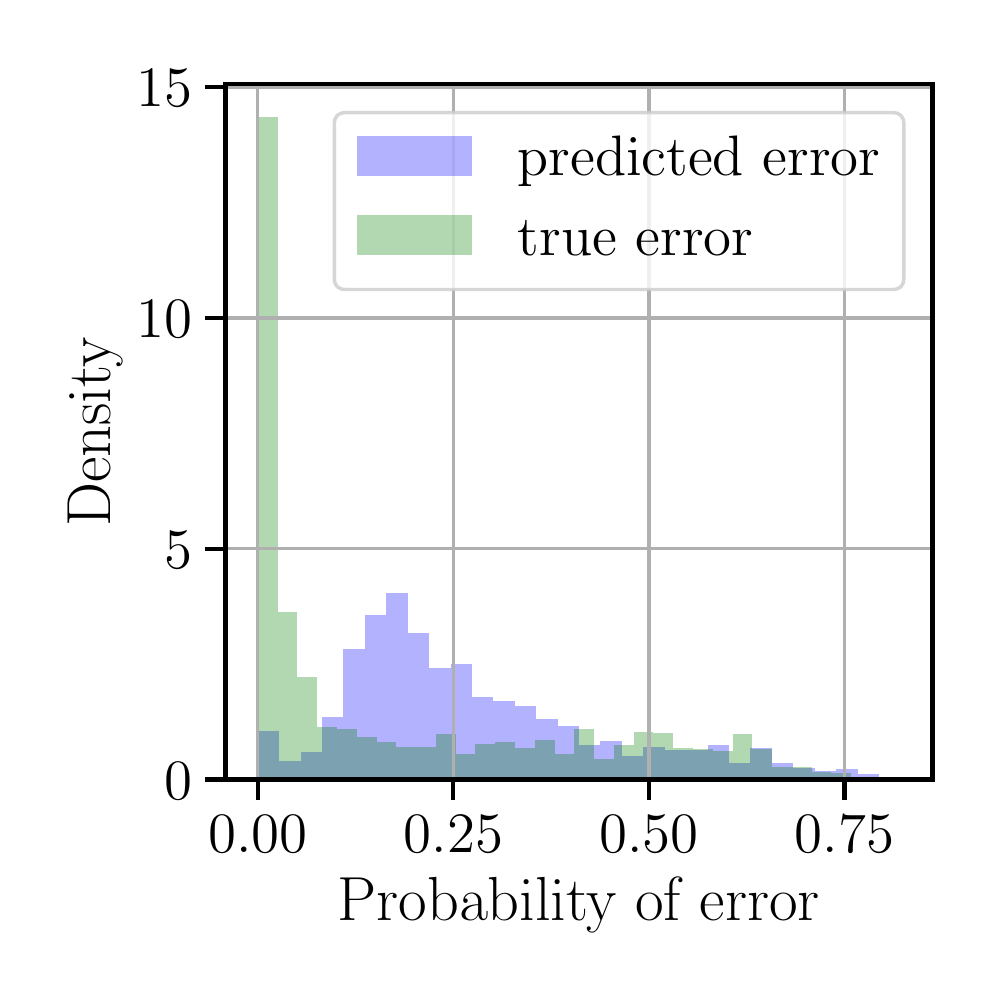}
         \caption{Risk for Model Trained with Softmax Surrogate}
         \label{fig:sontag_true_errror_pred_error_ham10000}
     \end{subfigure}
     \begin{subfigure}{0.45\textwidth}
         \centering
         \includegraphics[width=\linewidth, height=6.75cm]{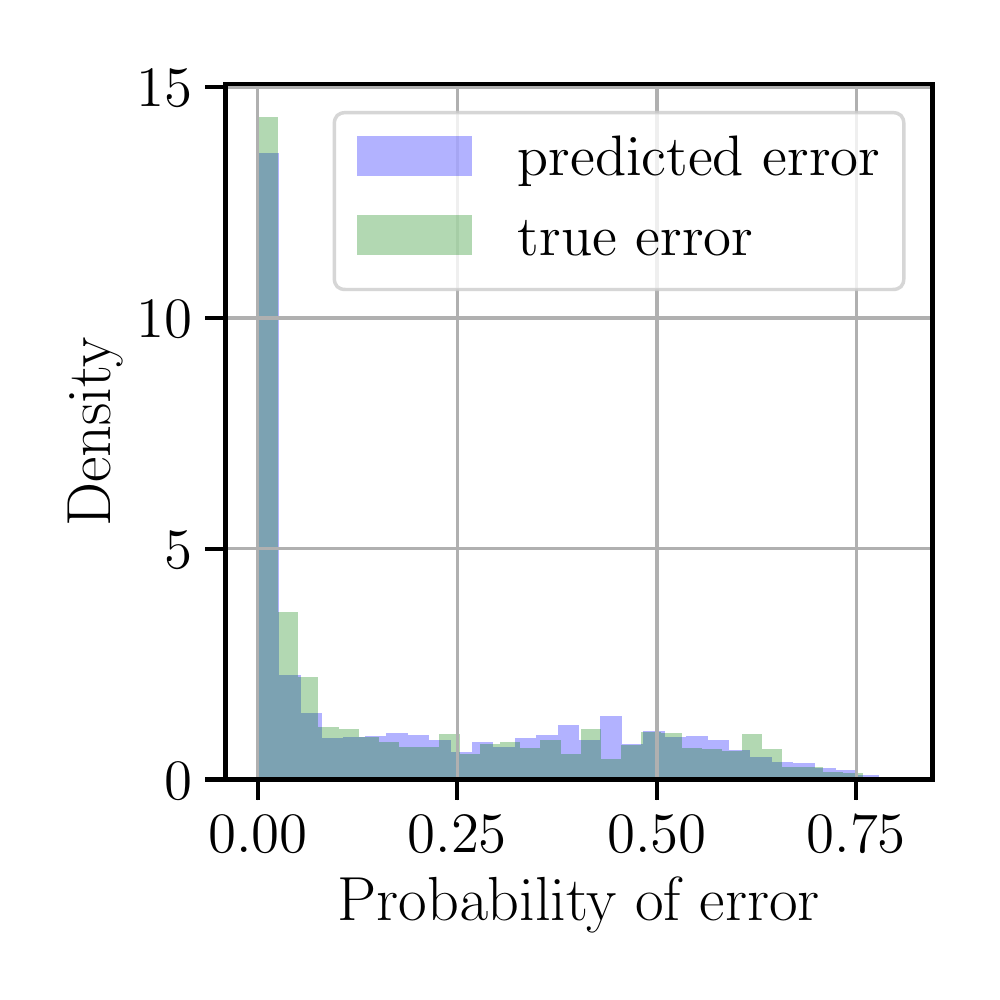}
         \caption{Risk for Model Trained with OvA Surrogate (Ours)}
         \label{fig:ova_true_error_pred_error_ham10000}
     \end{subfigure}
     \hfill
     \caption{\textit{Risk for Softmax vs OvA models on HAM10000}: Subfigure (a) reports the distribution of risks for the softmax method: $1-p_{\rsm}(\vx)$.  Subfigure (b) reports the distribution of risks for the OvA method: $1-p^{\text{\tiny{OvA}}}_{\rsm}(\vx)$.  We observe markedly more overlap for the latter.  The Wasserstein distance between the empirical and true error distributions is $8.02\pm1.37$ for OvA and  $26.72\pm1.77$ for softmax.}
     \label{fig:ham_error}
\end{figure*}

\section{Related Work}\label{sec:related}
Learning with a reject option (a.k.a.~rejection learning) is a long-studied problem, dating back to (at least) \citet{Chow1957AnOC}'s work on an optimal learning rule for a fixed rejection rate.  This initial work then stimulated a range of follow-up approaches, which can be categorized into two types: confidence-based \citep{10.5555/1390681.1442792, JMLR:v11:yuan10a, 10.5555/3327345.3327458, NIPS2008_3df1d4b9, Ramaswamy2018ConsistentAF, Ni2019OnTC} and classifier-rejector \citep{46544,NIPS2016_7634ea65}. The classifier-rejector approach has been well-studied for binary classification and resulted in theoretical guarantees \citep{46544, NIPS2016_7634ea65}. \citet{Ni2019OnTC} was the first to seriously study the multi-class formulation and found that the existing theory was hard to extend to this more general case. Most recently, \citet{pmlr-v139-charoenphakdee21a} proposed a surrogate loss for rejection learning for general classification, taking inspiration from cost-sensitive learning.


For safety-critical applications, rejection learning is a promising paradigm. However, its learning procedure completely ignores the downstream experts who will eventually make decisions for the rejected samples. \citet{10.5555/3327345.3327513} introduced an adaptive rejection framework termed \textit{learning to defer} (L2D).  L2D aims to directly model the interaction between the (usually human) decision makers and the autonomous system.  \citet{10.5555/3327345.3327513} propose a mixture of experts model for this end. \citet{raghu2019algorithmic} approaches the same problem by learning a classifier and comparing the expert's certainty and the classifier's certainty, deferring if the latter is lower. \citet{Wilder2020LearningTC} use the same mixture of experts framework as \citet{10.5555/3327345.3327513} and apply the same confidence-based deferral policy as \citet{raghu2019algorithmic}.  

In the work closest to ours, \citet{pmlr-v119-mozannar20b} study the L2D classification problem with generality, finding the algorithms proposed by \citet{10.5555/3327345.3327513} are inconsistent. They also study the limitation of confidence-based approaches \citep{raghu2019algorithmic}. Moreover, they propose the first consistent loss for multiclass L2D, establishing the importance of having a consistent surrogate. Our work, on the other hand, is the first to study the calibration of confidence estimates within L2D systems.




\section{Experiments}
We perform two types of experiments.  In the first, we verify that our OvA loss results in a better calibrated model for $\mathbb{P}(\rsm = \ry | \vx)$ than \citeauthor{pmlr-v119-mozannar20b}'s \citeyearpar{pmlr-v119-mozannar20b} loss.  We verify this in a CIFAR-10 simulation in Section \ref{sec:cifar10_exp}.  We then show that the softmax loss's mis-calibration has consequences for safety-critical decision making.  We train models for each loss on \texttt{HAM10000}, a data set for the diagnosis of skin lesions \citep{tschandl2018ham10000}, showing in Section \ref{sec:ham_exp} that our OvA model assesses risk more accurately than its softmax counterpart.

In the second type of experiment shown in Section \ref{sec:acc_exp}, we assess the overall accuracy on hate speech detection, galaxy classification, and skin lesion diagnosis.  We compare our OvA-based method to \citeauthor{pmlr-v119-mozannar20b}'s \citeyearpar{pmlr-v119-mozannar20b} as well as other state-of-the-art methods, such as differentiable triage \citep{okati2021differentiable}.  We find that our OvA models are at least competitive with, if not superior to, the best-performing competitor in all experiments.  Thus, our OvA method enjoys the benefits of calibration without any sacrifice to predictive performance. 

In all our implementations of \citeauthor{pmlr-v119-mozannar20b}'s \citeyearpar{pmlr-v119-mozannar20b} loss, we set the re-weighting parameter as $\alpha=1$.  Although \citet{pmlr-v119-mozannar20b} observe better performance when tuning $\alpha$, $\alpha=1$ is the \emph{only} value for which their surrogate is provably consistent.  The same is true for our loss and so our OvA surrogate does not include re-weighting either.  Comparing these losses in their `purest' forms is appropriate since our primary experimental concern is validating calibration.  For all OvA results, we use the logistic loss as the surrogate loss for binary classification.  Results are averaged over re-runs with six different random seeds.  Our software implementations are publicly available.\footnote{\href{https://github.com/rajevv/OvA-L2D}{https://github.com/rajevv/OvA-L2D}}

\subsection{Comparison to the Softmax Loss on \texttt{CIFAR-10}}\label{sec:cifar10_exp}

\paragraph{Data, Model, and Training} We use the standard train-test splits of \texttt{CIFAR-10} \citep{krizhevsky2009learning}.  We further partition the training split by $90\%-10\%$ to form training and validation sets, respectively.  We simulate the expert demonstrations from the training labels, as is described in detail below. We use the same neural network and training settings for both the OvA and softmax methods.  Following \citet{pmlr-v119-mozannar20b}, we use a wide residual networks \citep{zagoruyko2016wide} to parameterize the $g(\vx)$ functions. We train a $28$-layer network using stochastic gradient descent (SGD) with momentum and a cosine annealing schedule for the learning rate. We employ early stopping, terminating training if the validation loss does not improve for $20$ epochs. Additional experimental details can be found in Appendix \ref{sec:exp_details}. 

\paragraph{OvA Method's Calibration} We now test our OvA method's calibration in the same experimental setting used to test the softmax method in Section \ref{sec:problem_with_softmax_parameterization}.  To reiterate, the expert has a $75\%$ chance of being correct on the first five classes and random chance on the last five.  Figure \ref{fig:OvA_rejector_reliability} reports a reliability diagram and the ECE.  Comparing to the softmax results in Figure \ref{fig:sontag_rejector_reliability}, our OvA loss produces a model that has an over fifty percent reduction in ECE: $7.58\%$ for softmax, $3.01\%$ for OvA.  Figure \ref{fig:OvA_true_error_vs_predicted_error} reports the empirical distribution of error estimates: $1 - p^{\text{\tiny{OvA}}}_{\rsm}(\vx)$.  Unlike the corresponding softmax results in Figure \ref{fig:sontag_true_error_vs_predicted_error}, the OvA method produces sharper modes nearer to the true error values.  Moreover, OvA does not have a false mode at zero. 

\paragraph{Comparing Calibration Across Estimators}  We next test OvA's calibration against not only the softmax but also the proxy function $p_{\bot}$ from Equation \ref{eq:def_func}.  We consider two types of experts: a useful one and a random one. The useful one is an \textit{oracle} (i.e.~always correct) for the first seven classes and predicts randomly for the last three classes.  The random expert predicts uniformly over all classes.  Moreover, we consider when the data is useful, i.e.~the original CIFAR-10 training split, and when it is random, i.e.~training labels are uniformly random.  
\begin{table}[!h]
\centering
  \begin{tabular}{l c c c}
    \multicolumn{4}{c}{Expected Calibration Error ($\%$) on CIFAR-10} \\
       & \textbf{OvA} & Softmax & Proxy  \\
      \hline
      Both Random & $0.53$ & $0.97$ & $\mathbf{0.04}$  \\
      Random Expert & $\mathbf{0.68}$ & $3.72$ & $2.83$    \\
      Random Data & $\mathbf{2.05}$  & $2.07$ & $39.06$   \\ 
      Both Useful & $\mathbf{1.68}$ & $3.32$ & $37.15$   \\
  \end{tabular}
  \caption{\textit{ECE (\%) on CIFAR-10 Simulation.} We compare calibration across the three parameterizations considered: OvA (Eq.~\ref{eq:expert_prob_ova}), softmax (Eq.~\ref{eq:expert_prob}), and proxy (Eq.~\ref{eq:def_func}).}
  \label{tab:expertA_expertB_ECE}
  \end{table}

ECE results for the OvA (Eq.~\ref{eq:expert_prob_ova}), softmax (Eq.~\ref{eq:expert_prob}), and proxy (Eq.~\ref{eq:def_func}) methods are reported in Table \ref{tab:expertA_expertB_ECE}.  OvA has the best ECE in all but one case---the one in which both expert and data are random.  Yet the $p_{\bot}$ proxy is clearly not a viable estimator since it has an egregious ECE of $37.15\%$ when both data and expert are useful.  Furthermore, its ECE is an even worse $39.06\%$ when the expert is useful and data is random.  In general, the softmax's true estimator $p_{\rsm}$ is competent but still consistently worse than the OvA estimator. We compare the ECE values for the classifier for OvA and softmax in Table \ref{tab:ece_clf_values} in Appendix \ref{sec:ece_clf_values}.

\begin{figure*}
     \centering
     \begin{subfigure}[b]{\textwidth}
         \centering
         \includegraphics[width=\linewidth]{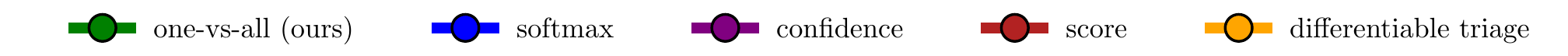}
     \end{subfigure}
     \hfill
       \begin{subfigure}[b]{0.312\textwidth}
         \centering
         \includegraphics[width=\linewidth, height=5.0cm]{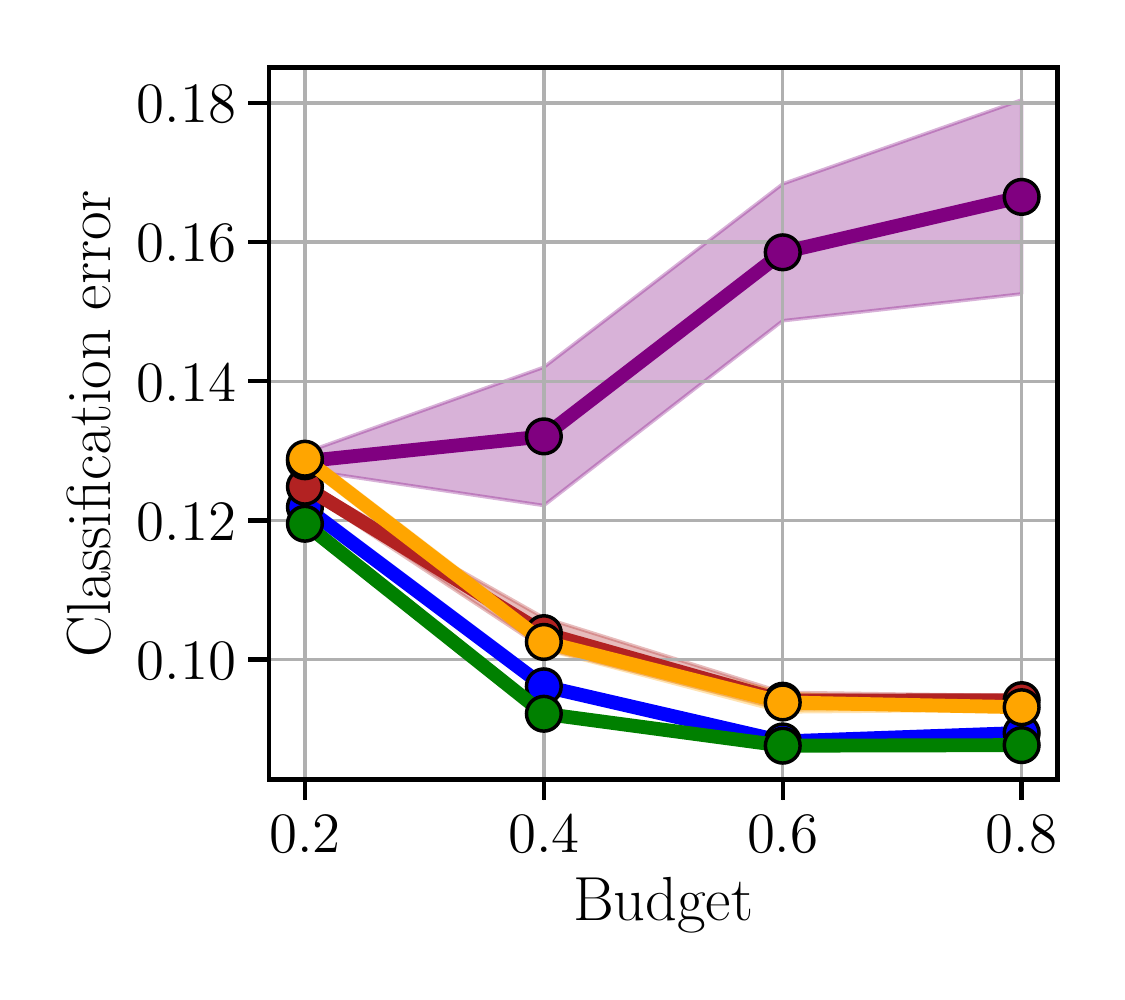}
         \caption{HateSpeech}
         \label{fig:misclassification_hatespeech}
     \end{subfigure}
     \hfill
     \begin{subfigure}[b]{0.312\textwidth}
         \centering
         \includegraphics[width=\linewidth, height=5.0cm]{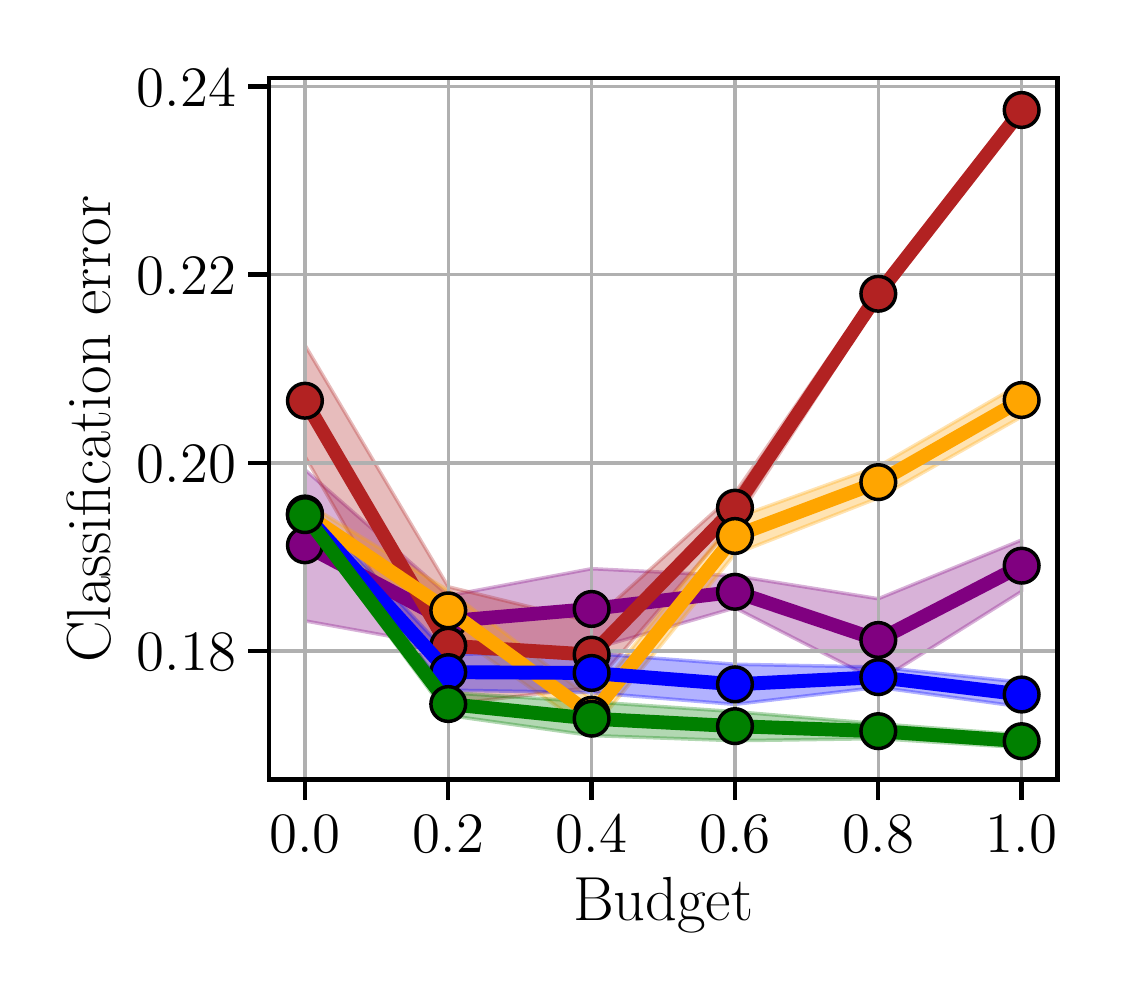}
         \caption{Galaxy-Zoo}
         \label{fig:misclassification_galaxy-zoo}
     \end{subfigure}
     \hfill
     \begin{subfigure}[b]{0.312\textwidth}
         \centering
         \includegraphics[width=\linewidth, height=5.0cm]{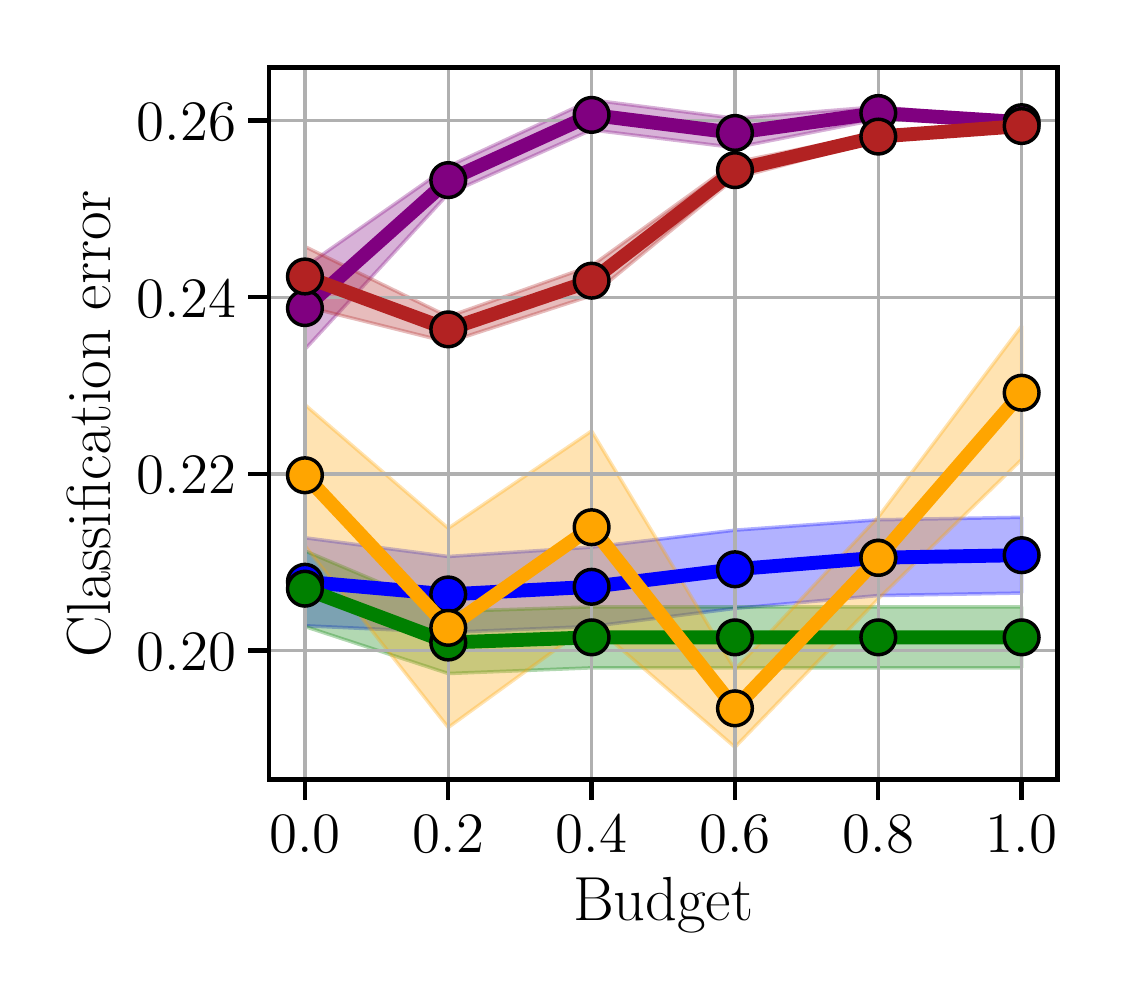}
         \caption{HAM10000}
         \label{fig:misclassification_ham10000}
     \end{subfigure}
     \caption{\textit{Accuracy on HateSpeech, Galaxy-Zoo, and HAM10000}: The subfigures report the classification error of OvA method, softmax method, and baselines for three data sets.  OvA (green) is competitive in all cases and is superior for \texttt{HateSpeech} and \texttt{Galaxy-Zoo}.}
     \label{fig:misclassification_error_all_methods}
\end{figure*}

\paragraph{System Accuracy and Coverage} For the final \texttt{CIFAR-10} experiment, we compare the OvA system's accuracy to the softmax's. The expert in this case has a $70\%$ chance of being correct if the image belongs to the classes $[1, k]$ and random chance if it belongs to classes $[k, 10]$.  We then vary $k$ from $k=2$ to $k=8$.  The left plot in Figure \ref{fig:coverage_accuracy_tradeoff} shows accuracy vs $k$.  Our OvA model (blue) has a modest but consistent advantage over the softmax model (red).  

The right plot in Figure \ref{fig:coverage_accuracy_tradeoff} reports the accuracy vs coverage, where coverage is the proportion of samples that the system has \emph{not} deferred.  \textit{Classifier accuracy} is the accuracy on the non-deferred samples. An L2D system ideally should have high coverage and high accuracy.  Again, the results show the OvA method's (blue) advantage at most coverage levels.  Note the OvA's significant superiority at low coverage ($0.2 - 0.3$).  Here the rejector must carefully choose which instances to pass to the classifier.  We conjecture that OvA's success is likely due to OvA's superior calibration in estimating when to defer.    



\subsection{Risk Assessment on \texttt{HAM10000}}\label{sec:ham_exp}

\paragraph{Data, Model, and Expert}  We again study risk assessment but this time for a high-stakes medical task.  \texttt{HAM10000} \citep{tschandl2018ham10000} is a data set of 10,015 dermatoscopic images containing seven categories of human skin lesions.  We partition the data into 60\% training, 20\% validation, and 20\% test splits. Each image includes metadata such as age, gender, and diagnosis type of the lesion. For our simulated expert model, we train an $8$-layer MLPMixer  \citep{Tolstikhin2021MLPMixerAA}. To simulate the expert having extra information, we input the image metadata into to the final feedforward layer. This model has a classification accuracy of 74\% (see Table \ref{tab:perf_mlp_mixer}). For the classifier, we fine-tune a 34-layer residual network (ResNet34) \citep{he2016deep}, following \citet{Tschandl2020HumancomputerCF}. We use data augmentations such as random cropping, reflection, and horizontal flipping.



\paragraph{Results} Figure \ref{fig:ham_error} visualizes the expert's predicted error and the expert's true error on the \texttt{HAM10000} test set.  Subfigure (a) shows results for the softmax method and (b) for our OvA method.  We restrict $p_{\rsm}(\vx) \in (0,1]$ for the softmax surrogate.  The gap between the predicted and true error is substantially reduced for OvA. We confirm this quantitatively by computing the Wasserstein distance between the true and predicted error. The distance is $8.02\pm1.37$ for OvA and $26.72\pm1.77$ for softmax. OvA provides clearly superior estimates of the expert's error.  

\subsection{Overall Accuracy}\label{sec:acc_exp}
\paragraph{Data} Lastly, we examine the OvA method's classification error on three real-world tasks: \texttt{HAM10000} \citep{tschandl2018ham10000}  for diagnosing skin lesions, \texttt{Galaxy-Zoo} \citep{10.1111/j.1365-2966.2008.14252.x} for scientific discovery, and \texttt{HateSpeech} \citep{Davidson2017AutomatedHS} for detecting offensive language.  Following \citet{okati2021differentiable}, we use a random sample of $10,000$ images for \texttt{Galaxy-Zoo}. We use $60\%$ train, $20\%$ validation, and $20\%$ test splits for \texttt{HAM10000} and \texttt{HateSpeech}.

\paragraph{Baselines} We compare the OvA- and softmax-based surrogates to three baselines.  The first is \textit{differentiable triage} \citep{okati2021differentiable}, a policy-learning method. The other two baselines are confidence-based methods that do not enjoy theoretical guarantees. The two are the \textit{score} baseline \cite{raghu2019algorithmic} and the \textit{confidence} baseline \cite{Bansal2021IsTM}. We give more details about the baselines and their implementation in Appendix \ref{sec:baselines}.

\paragraph{Models and Experts} We closely follow the setup of \citet{okati2021differentiable} for these experiments. Our base model is a $50$-layer residual network (ResNet50) for \texttt{Galaxy-Zoo}. For \texttt{HateSpeech}, we first embed the tweet's text into a $100$-dimensional feature vector using \textit{fasttext} \cite{DBLP:journals/corr/JoulinGBDJM16}. Our base model for \texttt{HateSpeech} is the text classification CNN developed by \citet{kim-2014-convolutional}. For the surrogate loss methods, we sample the expert demonstrations from the expert model's predictive distribution.  For training the surrogate models, we early stop if the validation loss does not improve for 20 epochs. We train the models using Adam \citep{DBLP:journals/corr/KingmaB14}, a cosine-annealed learning rate, and a warm-up period of 5 epochs. For other baselines, we use the same experimental setup as \citet{okati2021differentiable}. 


\paragraph{Results} Figure \ref{fig:misclassification_error_all_methods} reports the classification accuracy for each data set and each baseline as a function of the \textit{budget}.  The budget is the upper limit on the proportion of samples the system can defer to the expert.  The OvA surrogate is competitive among all baselines for the range of budgets considered. This shows that the OvA does not sacrifice accuracy for improved calibration. Rather, our model enjoys the benefits of both predictive performance and uncertainty quantification.  OvA's performance is also quite stable across random seeds.

\section{Conclusions}
In human-AI collaboration, it is vital that the system be reliable and trusted by the human.  Having a well-calibrated system---one that is a good forecaster---can help engender this trust. Our work investigates confidence calibration for the softmax surrogate loss, which previously was the only consistent surrogate for multiclass L2D.  We find that the softmax parameterization suffers from degenerate estimates of expert correctness. We solve this issue by deriving an alternative loss function that is also consistent. We experimentally show that our one-vs-all loss results in models better calibrated than those trained with the softmax-based surrogate.  In future work, we plan to investigate calibration in non-surrogate-based L2D systems, such as differentiable triage \citep{okati2021differentiable}.



\section*{Acknowledgements}
We thank Hussein Mozannar for helpful discussions and Daniel Barrejón for helping create the figures. This publication is part of the project \textit{Continual Learning under Human Guidance} (VI.Veni.212.203), which is financed by the Dutch Research Council (NWO).  This work was carried out on the Dutch national e-infrastructure with the support of SURF Cooperative.

\bibliography{references}
\bibliographystyle{icml2022}

\newpage
\appendix
\onecolumn

\section{A Primer on Calibration and Consistency for Classification}\label{sec:learning_theory_preliminaries}
\subsection{A General Classification Problem and Surrogate Losses}\label{sec:general_classification_problem}
Given $\mathcal{X} \subseteq \mathbb{R}^{n}$ as the input space, and $\mathcal{Y} = [n]$ as the output label space, we have an unknown distribution $\mathcal{D}$ over $\mathcal{X} \times \mathcal{Y}$. The output prediction label space is $\hat{\mathcal{Y}} = [k]$, and in general classification problem $k$ and $n$ can be different. The goal of the classification problem then is to learn a mapping $h: \mathcal{X} \rightarrow \hat{\mathcal{Y}}$. We assess the performance of the prediction function $h$ via a loss function $\ell: \mathcal{Y}\times \hat{\mathcal{Y}} \rightarrow \mathbb{R}_{+}$, and we aim to find $h$ with small $\ell$  \textit{-risk} which is defined as follows: 
\begin{align}
    \mathcal{R}_{\mathcal{D}}^{\ell}[h] = \mathbb{E}_{\vx, y \sim \mathcal{D}}\left[\ell\left(y, h\left(\vx \right)\right)\right]
\end{align}
We define the $\text{\textit{Bayes }} \ell$\textit{-risk} $\mathcal{R}_{\mathcal{D}}^{\ell, *}$ as the minimal $\ell$\textit{-risk} one can hope to achieve for the distribution $\mathcal{D}$, i.e $\mathcal{R}_{\mathcal{D}}^{\ell, *} := \inf_{h: \mathcal{X} \rightarrow \hat{\mathcal{Y}}}\mathcal{R}_{\mathcal{D}}^{\ell}[h]$. In practical settings , the classification learning problem assumes access to the training sample $\{\vx_{i}, y_{i}\}_{i=1}^{N}$ drawn independently and identically distributed from $\mathcal{D}$, and the learning algorithm seeks to learn $h$ by minimizing an empirical version of $\ell$ \textit{-risk} $\hat{\mathcal{R}}_{\mathcal{D}}^{\ell}[h]$. For $h \in \mathcal{H}$, $\hat{\mathcal{R}}_{\mathcal{D}}^{\ell}[h]$ is defined as
\begin{align}
    \hat{\mathcal{R}}_{\mathcal{D}}^{\ell}[h] = \frac{1}{N}\sum_{i=1}^{N}\ell\left(y_{i}, h\left(\vx_{i}\right)\right)
\end{align}

An important notion of success for such a learning algorithm is the convergence of $\mathcal{R}_{\mathcal{D}}^{\ell}[h_{S}] \rightarrow \mathcal{R}_{\mathcal{D}}^{\ell, *}$, i.e. when the learning algorithm receives increasingly large sample $S \sim \mathcal{D}^{N}$, the $\ell$-\text{\textit{risk}} of the function $h_{S}$  returned by the learning algorithm converges in probability to the \textit{Bayes} $\ell$\textit{-risk}, written formally as 
\begin{align}
    \forall \epsilon > 0 \; P_{S \sim \mathcal{D}^{N}}\left(\mathcal{R}_{\mathcal{D}}^{\ell}[h_{S}] > \mathcal{R}_{\mathcal{D}}^{\ell, *} + \epsilon \right) \rightarrow 0 \text{ as } N \rightarrow \infty
\end{align}

However, minimizing the $\ell$\textit{-risk} (similarly, empirical $\ell$\textit{-risk}) is computationally difficult for some classes of loss functions. For instance, for the misclassification loss $\ell_{0-1}: (y, \hat{y}) \mapsto \mathbb{I}(y \neq \hat{y})$, computationally minimizing $\ell-\text{\textit{risk}}$ is NP-hard. Thus, a surrogate loss $\psi(\cdot)$ over a surrogate prediction space $\mathcal{C} \subseteq \mathbb{R}^{k}$ is generally employed as a replacement for the target loss $\ell(\cdot)$.

For a surrogate prediction space $\mathcal{C} \subseteq \mathbb{R}^{k}$, a surrogate loss $\psi: \mathcal{Y} \times \mathcal{C} \rightarrow \mathbb{R}_{+}$, the goal is to learn a function $f: \mathcal{X} \rightarrow \mathcal{C}$ over some suitable class of functions $\mathcal{F}$, and a suitable decoding function $g: \mathcal{C} \rightarrow \hat{\mathcal{Y}}$. We then have the usual notions of $\mathcal{R}_{\mathcal{D}}^{\psi}[f]$ and $\mathcal{R}_{\mathcal{D}}^{\psi,*}$. An important question in such a setting is whether the convergence $\mathcal{R}_{\mathcal{D}}^{\psi}[f_{S}] \rightarrow \mathcal{R}_{\mathcal{D}}^{\psi, *}$ implies the convergence $\mathcal{R}_{\mathcal{D}}^{\ell}[g \circ f] \rightarrow \mathcal{R}_{\mathcal{D}}^{\ell,*}$. A positive answer to this question is necessary for the success of the classification problem learned by minimizing a surrogate loss $\psi(\cdot)$, and it is formally known as the \textit{consistency} of the surrogate loss $\psi(\cdot)$ w.r.t. the target loss $\ell(\cdot)$ as defined below:
\begin{definition}
($\mathcal{F}$-Consistency). A surrogate loss function $\psi(\cdot)$ is said to be $\mathcal{F}$-consistent with respect to the loss function $\ell(\cdot)$ if for any sequence of functions $f_{n} \in \mathcal{F}$
\begin{equation}
    \mathcal{R}_{\mathcal{D}}^{\psi}[f_{n}] \rightarrow \mathcal{R}_{\mathcal{D}}^{\psi, *} \implies \mathcal{R}_{\mathcal{D}}^{\ell}[g \circ f_{n}] \rightarrow \mathcal{R}_{\mathcal{D}}^{\ell, *}
\end{equation}
for all distributions $\mathcal{D}$.
\end{definition}

Define $\eta_{y}\left(\vx\right) = \mathbb{P}(\ry=y|\rvx=\vx)$ for each $y \in \mathcal{Y}$. $\vx$ and $y$ are the realizations of the random variables $\rvx$ and $\ry$ respectively over $\mathcal{X} \times \mathcal{Y}$. Then, we can write $\mathcal{R}_{\mathcal{D}}^{\ell}[h]$ as 
\begin{align}
    \mathcal{R}_{\mathcal{D}}^{\ell}[h] = \mathbb{E}_{\vx \sim \rvx}\left[\sum_{y=1}^{n}\eta_{y}\left(\vx\right)\ell\left(y, h\left(\vx\right)\right) \right] = \mathbb{E}_{\vx \sim \rvx}\left[\boldsymbol{\eta}\left(\vx\right)^{T}\boldsymbol{\ell}\left(h\left(\vx\right)\right)\right]
\end{align}
where $\boldsymbol{\eta}(\vx) = [\eta_{1}(\vx), \eta_{2}(\vx), \ldots, \eta_{n}(\vx)]^{T}$, and $\boldsymbol{\ell}(h(\vx)) = [\ell\left(\ry=1, h(\vx)\right), \ell\left(\ry, h(\vx)\right), \ldots, \ell\left(\ry=n, h(\vx)\right)]^{T}$. 
The quantity $\boldsymbol{\eta}(\vx)^{T}\boldsymbol{\ell}(h(\vx))$ is known as the
$\text{\textit{inner }} \ell-\text{\textit{risk}}$ denoted as $\mathcal{C}_{\boldsymbol{\eta(\vx)}, \vx}^{\ell}[h]$. More generally, $\forall \vx \in \mathcal{X}$, $\forall  \boldsymbol{\eta} \in [0,1]^{n}$, $\mathcal{C}_{\boldsymbol{\eta}, \vx}^{\ell}[h] := \boldsymbol{\eta}^{T}\boldsymbol{\ell}(\bold{h}(\vx))$ is known as the \textit{inner } $\ell$\textit{-risk}. We also define \textit{Bayes inner } $\ell${\textit{-risk} $\mathcal{C}_{\boldsymbol{\eta}, \vx}^{\ell, *} := \inf_{h:\mathcal{X} \rightarrow \hat{\mathcal{Y}}}\mathcal{C}_{\boldsymbol{\eta}, \vx}^{\ell}[h]$. We can also define these quantities for the surrogate loss $\psi(\cdot)$. A property called \textit{Calibration} of the \textit{inner } $\psi$\textit{-risk} of the surrogate loss $\psi(\cdot)$ w.r.t. \textit{inner } $\ell$\textit{-risk}} is then the necessary condition for the \textit{consistency} of the surrogate loss $\psi(\cdot)$ w.r.t $\ell(\cdot)$, and is usually a powerful tool for establishing and studying \textit{consistency} for surrogate losses. It is formally defined as follows:
\begin{definition}[{\citet{Steinwart2007HowTC}}]
($\mathcal{F}$-Calibration). A surrogate loss function $\psi(\cdot)$ is said to be $\mathcal{F}$-calibrated with respect to the loss function $\ell(\cdot)$ if, for all $\epsilon > 0$, $\boldsymbol{\eta} \in [0,1]^{n}$, and $\vx \in \mathcal{X}$, there exists $\delta > 0$ such that for any function $f \in \mathcal{F}$ 
\begin{equation}
    \mathcal{C}_{\boldsymbol{\eta}, \vx}^{\psi}[f] < \mathcal{C}_{\boldsymbol{\eta}, \vx}^{\psi, *} + \delta \implies \mathcal{C}_{\boldsymbol{\eta}, \vx}^{\ell}[g \circ f] < \mathcal{C}_{\boldsymbol{\eta}, \vx}^{\ell, *} + \epsilon.
\end{equation}
\end{definition}
As stated before, $\mathcal{F}$-calibration is a necessary condition for $\mathcal{F}$-consistency. However, with the satisfaction of an additional condition called \textit{minimizability} \cite{Steinwart2007HowTC}, $\mathcal{F}$-calibration also implies $\mathcal{F}$-consistency. We note that when $\mathcal{F} = \mathcal{F}_{\text{all}}$, i.e. when the hypothesis class consists of all measurable functions, \textit{minimizability} condition is satisfied. Thus, in such a case, it is enough to verify the calibration property of the surrogate loss to ensure the consistency of the surrogate loss w.r.t.  the target loss. Intuitively, a surrogate loss $\psi(\cdot)$ is said to be calibrated with respect to the target loss $\ell(\cdot)$ if minimizing $\psi(\cdot)$ results in a classifier $f$ with suitable decoding function $g$ whose \textit{inner } $\ell$\textit{-risk} is close to the \textit{Bayes inner } $\ell$\textit{-risk} for each $\vx \in \mathcal{X}$. Moreover, with an additional condition of \textit{minimizability}, calibration theoretically guarantees that for each $\vx \in \mathcal{X}$, the optimal solution of the \textit{inner} $\psi$-\textit{risk} minimization problem agrees with the optimal solution function of the $\ell$-\textit{risk} minimization problem evaluated at $\vx$. We state some important results for Binary Classification in the next section, and refer the reader to \citet{Steinwart2007HowTC} for more details.

\subsection{Calibration of Binary Surrogate Losses}\label{sec:binary_surrogate_loss_and_calibration}
Following the notation in the previous section, we have $\mathcal{Y} = \{-1, 1\}$. Here $\eta(\vx) = \mathbb{P}(Y=1|\rvx=\vx)$. We define the \textit{inner} $\ell$-\textit{risk} as $\mathcal{C}_{\eta, \vx}^{\ell}[h] = \eta \ell\left(1, h\left(\vx\right)\right) + \left(1 - \eta\right)\ell\left(-1, h\left(\vx\right)\right)$. Similarly, we define \textit{inner} $\psi$-\textit{risk} for a surrogate loss $\psi:\mathcal{Y}\times \mathcal{C} \rightarrow \mathbb{R}_{+}$ acting on a surrogate prediction space $\mathcal{C} \subseteq \mathbb{R}$. The calibration of binary surrogate losses (especially margin-based losses) with respect to the misclassification loss $\ell_{0-1}$ has been widely studied in the literature \cite{bartlett2006convexity}. In this section, we state some of the results in this direction.
\begin{definition}[{\citet{bartlett2006convexity}}]
For a surrogate prediction space $\mathcal{C} \subseteq \mathbb{R}$, we say a binary classification surrogate loss $\psi: \mathcal{Y} \times \mathcal{C} \rightarrow \mathbb{R}_{+}$ is classification-calibrated if, for any $\eta \neq \frac{1}{2}$, we have
\begin{align}
    \inf_{f\left(\vx\right)\left(\eta-\frac{1}{2}\right)} \mathcal{C}_{\eta, \vx}^{\psi}[f] > \inf_{f\left(\vx\right)} \mathcal{C}_{\eta, \vx}^{\psi}[f]
\end{align}
\end{definition}
The above definition states that minimizing a calibrated surrogate loss $\psi(\cdot)$ can give us the Bayes optimal binary classifier. It is well known that a convex $\psi(\cdot)$ is classification calibrated iff $\psi$ is differentiable in second argument at 0, and $\psi^{'}(\cdot, 0) < 0$ \cite{bartlett2006convexity}.

\subsection{Binary Proper Losses and Proper Composite Surrogate Losses}
In this section, we briefly review binary proper losses and proper composite surrogate losses. Recall the definition of $\mathcal{C}_{\eta, \vx}^{\psi}[f]$ from Section \ref{sec:binary_surrogate_loss_and_calibration} for some loss $\psi: \{-1, 1\} \times \mathcal{C} \rightarrow \mathbb{R}_{+}$. 
To simplify the notation, we rewrite $\mathcal{C}_{\eta, \vx}^{\psi}[f] = \mathcal{C}^{\psi}\left(\eta, f\left(\vx\right)\right) = \mathcal{C}^{\psi}\left(\eta, u\right)$ where $u=f\left(\vx\right)$. Next, we define \textit{proper composite losses}.



\begin{definition}[{\citet{pmlr-v35-ramaswamy14}}]
For $\mathcal{C} \subseteq \mathbb{R}$, a surrogate loss function $\psi:\{-1, 1\} \times \mathcal{C} \rightarrow \mathbb{R}_{+}$ is called proper composite loss if there exists a strictly increasing link function $\gamma: [0,1] \rightarrow \mathcal{C}$ such that:
\[\gamma\left(p\right) \in \argmin_{u \in \mathcal{C}}\mathcal{C}^{\psi}\left(p, u\right), \forall\; p \in [0,1]\]
If the above minimizer is unique for all $p \in [0,1]$, then we call the surrogate loss strictly proper composite loss.
\end{definition}


An important property of strictly proper composite losses is that their minimization leads to Fisher consistent class probability estimates \cite{Buja2005LossFF}. Logistic Loss ($\psi\left(y,u\right) = \log\left(1 + \exp\left(-yu\right)\right)$ is a common example of a strictly composite proper composite loss with the inverse link function $\gamma^{-1} = \frac{1}{1 + \exp\left(-u\right)}$. Thus, 
if $f: \mathcal{X} \rightarrow \mathbb{R}$ is learnt by minimizing the logistic loss, then $\hat{p}_{\vx} = \frac{1}{1 + \exp({-f(\vx)})}$ acts as a class probability estimate. Furthermore, strictly proper composite losses are classification calibrated \cite{JMLR:v11:reid10a}. Thus, based on the definition of calibration of the binary surrogate losses(Section \ref{sec:binary_surrogate_loss_and_calibration}), we can write the final predictor learnt by minimizing the logistic loss as:
\begin{equation}\label{eqn:composite_proper_decision}
h(\vx) = \sign(f(\vx)) = \sign(\hat{p}(\vx) - \frac{1}{2})
\end{equation}

\subsection{Code Based Surrogates for Multiclass Classification}\label{sec:code_matrix_method_for_classification}
Code Based methods are a class of classification techniques where some code matrix is used to decompose the multiclass classification problem into multiple binary classification problems. Mention could be made of error-correcting coding mechanism \citep{10.5555/1622826.1622834, Langford05sensitiveerror, 10.1162/15324430152733133}. We briefly describe the setup here, and refer the reader to \citet{pmlr-v35-ramaswamy14} for full details. The goal of such a code based mechanism it to use a code matrix $\bold{M} = \{\pm 1, 0\}^{n \times k}$ to decompose a $n$-class classification problem into $k$ binary classification problems. Following the notation from Section \ref{sec:general_classification_problem}, we use $\bold{M}$ to split the training sample $S = \{\left(\vx_{i}, y_{i}\right)\}_{i=1}^{N}$ into $k$-training samples $\tilde{S}_{j}$ for each $j \in [k]$ such that $\tilde{S}_{j} = \{\left(\vx_{i}, M_{y_{i},j}\right); i\in [1,N], M_{y_{i}, j} \neq 0\}$. Thus, each $\tilde{S}_{j}$ is a subset from the original $S$ with output (binary)labels replaced provided by the $\bold{M}$. For $\mathcal{C} \subseteq \mathbb{R}$, we use these $\tilde{S}_{j}$ to learn a $k$-binary classifiers $f_{j}: \mathcal{X} \rightarrow \mathcal{C}$. Thus, for each $\vx \in \mathcal{X}$, we get a prediction $f(\vx) = \left[f_{1}\left(\vx\right), \ldots, f_{k}\left(\vx\right)\right] \in \mathbb{R}^{k}$. We then a use a suitable decoding function to map $f\left(\vx\right)$ to the original prediction space $\hat{\mathcal{Y}}$.
If we use some suitable surrogate loss $\ell: \{-1,1\} \times \mathcal{C} \rightarrow \mathbb{R}_{+}$, then intuitively, the whole code matrix based mechanism can be viewed as learning a function $f: \mathcal{X} \rightarrow \mathcal{C}^{k}$ by minimizing a surrogate multiclass classification loss $\psi: \mathcal{Y} \times \mathcal{C}^{k} \rightarrow \mathbb{R}_{+}$ given as 
\begin{align}\label{eq:code_matrix_general_surrogate_loss}
\psi\left(\bold{y}, \bold{u}\right) = \sum_{j=1}^{k}\left(\mathbb{I}\left(M_{yj}=1\right)\ell\left(1, u_{j}\right) + \mathbb{I}\left(M_{yj} = -1\right)\ell\left(-1, u_{j}\right)\right)
\end{align}
Obviously, we care about the consistency of such a surrogate loss $\psi(\cdot)$ for a successful classification algorithm. \citet{pmlr-v35-ramaswamy14} analyze the conditions related to consistency and calibration of such a surrogate loss for general losses.

\section{One-vs-All surrogate Loss for L2D}\label{sec:OvA_closed_form_expression}
We derive the closed-form expression for surrogate loss $\rpsi_{\text{OvA}}$ using the procedure described in Appendix \ref{sec:code_matrix_method_for_classification} for the code matrix $\bold{M}$ defined Section \ref{sec:OvA_surrogate_loss_for_L2D}. Following the notation from Appendix \ref{sec:code_matrix_method_for_classification}, we have $n=K$ and $k=K+1$ for our L2D problem. For the surrogate prediction space $\mathbb{R}$, and $g_{y}: \mathcal{X} \rightarrow \mathbb{R}, y \in \mathcal{Y}$ and $g_{\bot}: \mathcal{X} \rightarrow \mathbb{R}$ and $\boldsymbol{g}(\rvx) = [g_{1}(\rvx), \ldots, g_{\bot}(\rvx)]$, we can use $\bold{M}$ to derive the closed form expression for the surrogate loss $\rpsi: \mathcal{Y} \times \mathbb{R}^{n+1} \rightarrow \mathbb{R}$ as follows:
\begin{enumerate}
    \item \textbf{Case 1:} $\rpsi\left(\boldsymbol{g}; \boldsymbol{x}, y, m \right)$ for $y$ such that $\mathbb{I}\left[y \neq \rsm \right] = 1$ \newline
    In this case, we can follow the definition of $\bold{M}$ to gather that $m_{yj} = 1$ only if $j=y$. Thus, we can follow Eqn. \ref{eq:code_matrix_general_surrogate_loss}, and get
    \[\rpsi\left(\boldsymbol{g}; \boldsymbol{x}, y, m\right) = \rphi\left[g_{y}\left(\boldsymbol{x}\right)\right] \;+ \sum_{\substack{y^{'} \in \mathcal{Y}\cup\{\bot\} \\ y^{'} \neq y}}\rphi\left[-g_{y^{'}}\left(\boldsymbol{x}\right)\right]\]
    
    \item \textbf{Case 2:} $\rpsi\left(\boldsymbol{g}; \boldsymbol{x}, y, m\right)$ for $y$ such that $\mathbb{I}\left[y = m\right] = 1$ \newline
    In this case, we have $m_{yy} = 1$ as well as $m_{y\bot} = 1$ where $\bot$ denotes the index $(K+1)$. Thus,
    \[\rpsi\left(\boldsymbol{g}; \boldsymbol{x}, y, m\right) = \rphi\left[g_{y}\left(\boldsymbol{x}\right)\right] + \rphi\left[g_{\bot}\left(\boldsymbol{x}\right)\right] \;+ \sum_{y^{'} \in \mathcal{Y},  y^{'} \neq y}\rphi\left[-g_{y^{'}}\left(\boldsymbol{x}\right)\right]\]
    
\end{enumerate}

Finally, we can combine both the cases to get
\[\rpsi\left(\boldsymbol{g}; \boldsymbol{x}, y, m\right) = \rphi\left[g_{y}\left(\boldsymbol{x}\right)\right] + \rphi\left[-g_{\bot}\left(\boldsymbol{x}\right)\right] + \sum_{y^{'} \in \mathcal{Y}, y^{'} \neq y}\rphi\left[-g_{y^{'}}\left(\boldsymbol{x}\right)\right] + \mathbb{I}\left[m=y\right]\left(\rphi\left[g_{\bot}\left(\boldsymbol{x}\right)\right] - \rphi\left[-g_{\bot}\left(\boldsymbol{x}\right)\right]\right)\]

where $\rphi: \{\pm 1\} \times \mathbb{R} \rightarrow \mathbb{R}_{+}$ is a binary classification surrogate loss, and $\rphi\left[g_{y}\left(\boldsymbol{x}\right)\right] = \rphi(1, g_{y}\left(\boldsymbol{x}\right))$. Similarly, $\rphi\left[-g_{y}\left(\boldsymbol{x}\right)\right] = \rphi(-1, g_{y}\left(\boldsymbol{x}\right))$.

\section{Proofs}

\subsection{Derivation of $p_{\rsm}\left(\vx\right)$ and $p_{k}\left(\vx\right)$ for the Softmax Surrogate Loss} \label{sec:probs_from_softmax_outputs}
Let $\mathcal{Y^{\bot}} = \mathcal{Y} \cup \{\bot\}$.  From the proof of Theorem 1 of \citet{pmlr-v119-mozannar20b}, we have that for the (Bayes) optimal $g^{*}_{1}, \ldots, g^{*}_{K}, g^{*}_{\bot}$:
\begin{equation}
    \begin{split}
        \frac{\mathbb{P}(\rsm = \ry | \vx)}{1 + \mathbb{P}(\rsm = \ry | \vx)} &=  \frac{\exp g^{*}_{\bot}(\vx)}{\sum_{y' \in \mathcal{Y}^{\bot}}\exp g^{*}_{y'}(\vx)}
        \\ &= p_{\bot}^{*}(\vx) 
    \end{split}
\end{equation} where $p^{*}_{\bot}(\vx)$ is the function we define in Equation \ref{eq:def_func} evaluated at the Bayes optimal $g$'s.
Rearranging, we then have: \begin{equation}
    p^{*}_{\bot}(\vx) = \frac{\mathbb{P}(\rsm = \ry | \vx)}{1 + \mathbb{P}(\rsm = \ry | \vx)} = \frac{1}{\mathbb{P}^{-1}(\rsm = \ry | \vx) + 1}.
\end{equation}  Solving for $\mathbb{P}(\rsm = \ry | \vx)$, we have: \begin{equation}
    \begin{split}
    \mathbb{P}(\rsm = \ry | \vx) &= \frac{1}{(p_{\bot}^{*}(\vx))^{-1} -1} \\
    &= \frac{p_{\bot}^{*}(\vx)}{1 - p_{\bot}^{*}(\vx)}
\end{split}
\end{equation}
Similarly, from the proof of Theorem 1 of \citet{pmlr-v119-mozannar20b}, we have for the Bayes Optimal $g_{k}^{*}$, $k \in [K]$:
\begin{equation}
    \begin{split}
    \frac{\mathbb{P}\left(\ry = k\vert\vx\right)}{1 + \mathbb{P}\left(\rsm = \ry| \vx\right)} &= \frac{\exp g^{*}_{k}(\vx)}{\sum_{y' \in \mathcal{Y}^{\bot}}\exp g^{*}_{y'}(\vx)}
    \end{split}
\end{equation}

\begin{equation}\label{eq:softmax_clf_prob}
    \begin{split}
        &\implies p_{k}(\vx) = \mathbb{P}(\ry = k\vert \vx) = \frac{1}{1 - p^{*}_{\bot}(\vx)}\frac{\exp g^{*}_{k}(\vx)}{\sum_{y' \in \mathcal{Y}^{\bot}}\exp g^{*}_{y'}(\vx)}
    \end{split}
\end{equation}

\subsection{Proof of Theorem \ref{thm: OvA_loss_calibration}} \label{sec:OvA_surrogate_loss_is_calibrated}

For $K+1$ surrogate prediction function $g_{1}\left(\rvx\right), \ldots, g_{K}\left(\rvx\right), g_{\bot}\left(\rvx\right)$, and the binary classification surrogate $\rphi : \{\pm 1\} \times \mathbb{R} \rightarrow \mathbb{R}_{+}$, the proposed one-vs-all (OvA) surrogate is has the following point-wise form:

\begin{equation}\begin{split}
     \rpsi_{\text{OvA}}&(g_{1},\ldots, g_{K}, g_{\bot}; \vx, y, m) = \\ & \rphi[g_{y}(\vx)] \ + \  \rphi[-g_{\bot}(\vx)] 
      \ +  \sum_{y' \in \mathcal{Y}, y' \ne y} \rphi[-g_{y'}(\vx)]  \mathbb{I}[m=y]\left(\rphi[g_{\bot}(\vx)] - \rphi[-g_{\bot}(\vx)]\right)
\end{split}
\end{equation}

We consider the point-wise \textit{inner} $\rpsi$\textit{-risk} for some $\rvx = \vx$ written as follows:
\begin{equation}
    \mathbb{E}_{\ry|\rvx = \vx}\mathbb{E}_{\rsm |\rvx=\vx, y}\rpsi_{\text{OvA}}(g_{1},\ldots, g_{K}, g_{\bot}; \vx, y, m)
\end{equation}

We simplify the \textit{inner} $\rpsi$\textit{-risk} by expanding both the expectations below:

\begin{equation} \begin{split}
 \mathbb{E}_{\ry|\rvx = \vx}&\mathbb{E}_{\rsm |\rvx=\vx, y}\rpsi_{\text{OvA}}(g_{1},\ldots, g_{K}, g_{\bot}; \vx, y, m) = \\ & \mathbb{E}_{\ry|\rvx = \vx}\bigg[\rphi(g_{y}(\vx)) + \rphi(-g_{\bot}(\vx)) + \sum_{y^{'} \in \mathcal{Y}, y^{'} \neq y} \rphi(-g_{y^{'}}(\vx)) \\ &+ \sum_{m \in \mathcal{Y}}\mathbb{P}(\rsm=m|\rvx=\vx, \ry=y)\mathbb{I}\left[m=y\right]\left[\rphi(g_{\bot}(\vx)) - \rphi(-g_{\bot}(\vx))\right]\bigg]
\end{split}
\end{equation}

Expanding the outer expectation, and $\eta_{y}\left(\vx\right) = p\left(\ry=y|\rvx=\vx\right)$

\begin{equation*} \begin{split}
 \mathbb{E}_{\ry|\rvx = \vx}&\mathbb{E}_{\rsm |\rvx=\vx, y}\rpsi_{\text{OvA}}(g_{1},\ldots, g_{K}, g_{\bot}; \vx, y, m) = \\ & \sum_{y \in \mathcal{Y}}\eta_{y}\left(\vx\right)\bigg[\rphi\left(g_{y}\left(\vx\right)\right) + \sum_{y^{'} \in \mathcal{Y}, y^{'} \neq y}\rphi\left(-g_{y^{'}}\left(\vx\right)\right)\bigg]  + \rphi\left(-g_{\bot}\left(\vx\right)\right)  \\ &+  \ \  \sum_{y \in \mathcal{Y}}\eta_{y}\left(\vx\right)\sum_{m \in \mathcal{Y}} \mathbb{P}\left(\rsm=m|\rvx=\vx, \ry=y\right)\mathbb{I}\left[m=y\right]\left[\rphi\left(g_{\bot}\left(\vx\right)\right) - \rphi\left(-g_{\bot}\left(\vx\right)\right)\right]
  \\ &= \ \ \sum_{y \in \mathcal{Y}}\eta_{y}\left(\vx\right)\bigg[\rphi\left(g_{y}\left(\vx\right)\right) + \sum_{y^{'} \in \mathcal{Y}, y^{'} \neq y}\rphi\left(-g_{y^{'}}\left(\vx\right)\right)\bigg]  + \rphi\left(-g_{\bot}\left(\vx\right)\right)   \\ &+  \ \  \sum_{y \in \mathcal{Y}}\eta_{y}\left(\vx\right)\sum_{m \in \mathcal{Y}} \mathbb{P}\left(\rsm=y|\rvx=\vx, \ry=y\right)\left[\rphi\left(g_{\bot}\left(\vx\right)\right) - \rphi\left(-g_{\bot}\left(\vx\right)\right)\right]
 \\ &= \ \ \sum_{y \in \mathcal{Y}}\eta_{y}\left(\vx\right)\bigg[\rphi\left(g_{y}\left(\vx\right)\right) + \sum_{y^{'} \in \mathcal{Y}, y^{'} \neq y}\rphi\left(-g_{y^{'}}\left(\vx\right)\right)\bigg]  + \rphi\left(-g_{\bot}\left(\vx\right)\right) \\ &+  \ \  \underbrace{\sum_{y \in \mathcal{Y}}\eta_{y}\left(\vx\right)\sum_{m \in \mathcal{Y}} \mathbb{P}\left(\rsm=y|\rvx=\vx, \ry=y\right)}_{\text{$\mathbb{P}(\ry = \rsm|\rvx = \vx)$}}\left[\rphi\left(g_{\bot}\left(\vx\right)\right) - \rphi\left(-g_{\bot}\left(\vx\right)\right)\right]
 \\ &= \ \ \sum_{y \in \mathcal{Y}}\eta_{y}\left(\vx\right)\bigg[\rphi\left(g_{y}\left(\vx\right)\right) + \sum_{y^{'} \in \mathcal{Y}, y^{'} \neq y}\rphi\left(-g_{y^{'}}\left(\vx\right)\right)\bigg]  + \rphi\left(-g_{\bot}\left(\vx\right)\right) \\ &+ \ \  \mathbb{P}(\ry = \rsm|\rvx = \vx)\left[\rphi\left(g_{\bot}\left(\vx\right)\right) - \rphi\left(-g_{\bot}\left(\vx\right)\right)\right]
 \\ &= \ \ \sum_{y \in \mathcal{Y}}\eta_{y}\left(\vx\right)\bigg[\rphi\left(g_{y}\left(\vx\right)\right) + \sum_{y^{'} \in \mathcal{Y}, y^{'} \neq y}\rphi\left(-g_{y^{'}}\left(\vx\right)\right)\bigg]   +  \mathbb{P}(\ry = \rsm|\rvx = \vx)\rphi\left(g_{\bot}\left(\vx\right)\right)  \\ &+ \ \ \left(1 - \mathbb{P}\left(\ry = \rsm|\rvx=\vx\right)\right)\rphi\left(-g_{\bot}\left(\vx\right)\right)
    \end{split}
\end{equation*}

Using the usual notation $p_{\rsm}(\vx) = p(\ry=\rsm|\rvx=\vx)$, we can further rewrite the above equation in the following form,

\begin{equation}\begin{split}
\mathbb{E}_{\ry|\rvx = \vx}&\mathbb{E}_{\rsm |\rvx=\vx, y}\rpsi_{\text{OvA}}(g_{1},\ldots, g_{K}, g_{\bot}; \vx, y, m) = \\ &
\sum_{y \in \mathcal{Y}}\left[\eta_{y}\left(\vx\right)\rphi\left(g_{y}\left(\vx\right)\right) + \left(1 - \eta_{y}\left(\vx\right)\right)\rphi\left(-g_{y}\left(\vx\right)\right)\right] + p_{\rsm}\left(\vx\right)\rphi\left({g_{\bot}\left(\vx\right)}\right) + \left(1 - p_{\rsm}\right)\rphi\left(-g_{\bot}\left(\vx\right)\right)
 \end{split}   
\end{equation}


The above expression says that we have $K+1$ binary classification problems where the \textit{inner} $\rphi$\textit{-risk} for the $i^{th}$ binary classification problem is given as $\eta_{y}\left(\vx\right) \rphi\left(g_{y}\left(\vx\right)\right) + \left(1 - \eta_{y}\left(\vx\right)\right)\rphi\left(-g_{y}\left(\vx\right)\right)$ when $i \in [K]$ and $p_{\rsm}\left(\vx\right)\rphi\left(g_{\bot}\left(\vx\right)\right) + \left(1 - p_{\rsm}\left(\vx\right)\right) \rphi\left(-g_{\bot}\left(x\right)\right)$ when $i \in \{K+1\}$. This means that the point-wise minimizer of the inner $\rpsi$\textit{-risk} can be analyzed in terms of the point-wise minimizer of the \textit{inner} $\rphi$\textit{-risk} for each of the $K+1$ binary classification problems we have. Denote the minimizer of point-wise \textit{inner} $\rpsi_{\text{OvA}}$\textit{-risk} as $\boldsymbol{g}^{*}$, then the above decomposition means $g^{*}_{i}$ corresponds to the minimizer of the \textit{inner} $\rphi$\textit{-risk} for the $i^{th}$ binary classification problem. 

We know that the Bayes solution for the binary classification problem is $\sign\left(\eta(\vx) - \frac{1}{2}\right)$ where $\eta(\vx)$ denotes $p(\ry=1|\rvx=\vx)$. Now when the binary surrogate loss $\rphi$ is a strictly proper composite loss for binary classification, by the property of strictly proper composite losses, we have $\sign(g^{*}_{y}(\vx))$ would agree with the Bayes solution of the Binary classification (refer Eqn. \ref{eqn:composite_proper_decision}), i.e. $g^{*}_{y}(\vx) > 0$ if $\eta_{y}\left(\vx\right) > \frac{1}{2}$. And similarly $g^{*}_{\bot}\left(\vx\right) \geq 0$ if $p_{\rsm}\left(\vx\right) > \frac{1}{2}$. Furthermore, we have the existence of a continuous and increasing inverse link function $\rgamma^{-1}$ for the binary surrogate $\rphi$ with the property that $\rgamma^{-1}\left(g^{*}_{y}\left(\vx\right)\right)$ would converge to $\eta_{y}\left(\vx\right)$. Similarly, $\rgamma^{-1}\left(g^{*}_{\bot}\left(\vx\right)\right)$ would converge to $p_{\rsm}\left(\vx\right)$.

Using the above, we can establish the Bayes optimal decision for this minimizer $\boldsymbol{g}^{*}$ using following cases.

\paragraph{Case 1:} If we have $g^{*}_{y}\left(\vx\right) > 0$ and $g^{*}_{\bot}\left(\vx\right) > 0$ for some $y \in \mathcal{Y}$. Note that we cannot have $y \neq y^{'}$ both belonging to $[K]$ such that $g^{*}_{y}(\vx) > 0$ and $g^{*}_{y^{'}}(\vx) > 0$. Because this would imply $\eta_{y}(\vx) > \frac{1}{2}$ and $\eta_{y^{'}}(\vx) > \frac{1}{2}$ which contradicts the rules of probabilities. Thus, theoretically, only one such $y \in \mathcal{Y}$ is possible such that $g^{*}_{y}\left(\vx\right) > 0 $. And if we take the prediction for our L2D problem as $\argmax_{k \in [K+1]}g^{*}_{k}\left(\vx\right)$, our prediction would correspond to the Bayes Optimal decision, i.e. if 
\[g^{*}_{y}(\vx) < g^{*}_{\bot}\left(\vx\right) \;\; \forall y \in \mathcal{Y}\]
\[\implies \rgamma^{-1}\left(g^{*}_{y}\left(\vx\right)\right) < \rgamma^{-1}\left(g^{*}_{\bot}\left(\vx\right)\right)\;\; \forall y \in \mathcal{Y}\]
\[\implies \eta_{y}\left(\vx\right) < p_{\rsm}\left(\vx\right) \;\; \forall y \in \mathcal{Y}\]
Thus, such if $g^{*}_{\bot}(\vx) > g^{*}_{y}(\vx)$ such that $g^{*}_{\bot}(\vx)  > 0, g^{*}_{y}(\vx) > 0$, then the prediction following the decision rule $\argmax_{k \in [K+1]}g^{*}_{k}\left(\vx\right)$ would correspond with the Bayes optimal rule
\[r\left(\vx\right) =  \mathbb{I}\left[\max_{\eta_{y \in \mathcal{Y}}}\eta_{y}\left(\vx\right) < p_{\rsm}\left(\vx\right)\right]\]

\paragraph{Case 2:} In this case, if $\nexists y \in \mathcal{Y}$ s.t. $g^{*}_{y}\left(\vx\right) > 0$, but $g^{*}_{\bot}(\vx) > 0$, then the same argument as above implies the decision with the Bayes optimal rule.

\paragraph{Case 3:} if $\exists y \in \mathcal{Y}$ s.t. $g^{*}_{y}\left(\vx\right) > 0$, but $g^{*}_{\bot}(\vx) < 0$, then the same argument as above implies the decision with the Bayes optimal rule. In this case, we will have $r(\vx) = 0$, and the classifier's prediction would correspond with the regular Bayes Optimal Classifier, i.e. $\argmax_{y \in \mathcal{Y}}\eta_{y}(\vx)$.

\paragraph{Case 4:} In this case, if $\nexists y \in \mathcal{Y}$ s.t. $g^{*}_{y}\left(\vx\right) > 0$, and also $g^{*}_{\bot}(\vx) < 0$. This situation invokes the common ``None of the above" classification rule for One-vs-All classifiers.

Thus, the cases above imply that the minimizer of the point-wise \textit{inner} $\rpsi$\textit{-risk} gives the Bayes Optimal Classifier and Rejection prediction for $\rvx = \vx$. Thus, the surrogate loss $\rphi$ is calibrated for 0-1 L2D.

\section{Additional Results}\label{sec:additional_results_appendix}

\subsection{ECE values with respect to the classifier correctness}\label{sec:ece_clf_values}
\begin{table}[!h]
\centering
  \begin{tabular}{l c c}
    \multicolumn{3}{c}{Expected Calibration Error ($\%$) on CIFAR-10} \\
       & \textbf{OvA} & Softmax  \\
      \hline
      Both Random & $0.51$ & $\mathbf{0.34}$   \\
      Random Expert & $\mathbf{6.47}$ & $7.22$    \\
      Random Data & $\mathbf{1.94}$  & $2.36$   \\ 
      Both Useful & $\mathbf{6.92}$ & $7.92$    \\
  \end{tabular}
  \caption{\textit{ECE for Classifier on CIFAR-10 Simulation.} We compare calibration across the two parameterizations: OvA (Eq.~\ref{eq:OvA_clf_prob}) and softmax (Eq.~\ref{eq:softmax_clf_prob}).}
  \label{tab:ece_clf_values}
  
  \end{table}

\subsection{Effect of Calibration on System's Accuracy}
\begin{figure*}[!h]
     \vskip 0.2in
         \centering
         \includegraphics[width= 6.25cm, height=6.275cm]{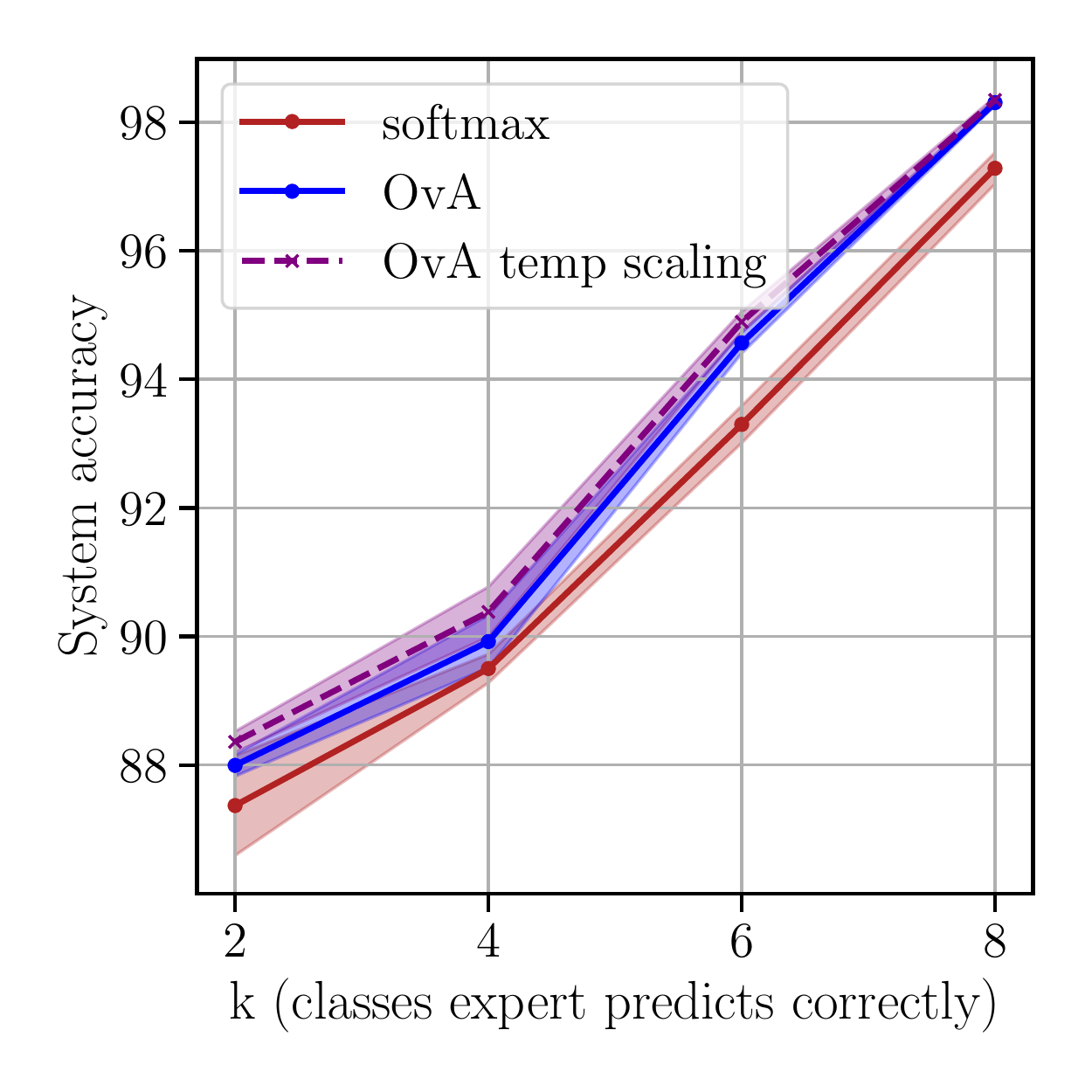}
         \caption{\textit{Effect of post-processing calibration for One-Vs-All rejector.} We can see that post-processing calibration of $p_{\rsm}(\vx)$ further improves the system accuracy. This shows the effect of calibration for the overall system's accuracy for L2D.}
         \label{fig:temp_scaling_ova_rej}
        \vskip -0.2in
\end{figure*}

In this section, we verify calibration's role in the overall system's accuracy. For the trained one-vs-all model from Figure \ref{fig:coverage_accuracy_tradeoff}, we apply a post-processing calibration technique called \textit{temperature scaling} \citep{10.5555/3305381.3305518} to further calibrate the rejector. In Figure \ref{fig:temp_scaling_ova_rej}, we see that this additional calibration step marginally improves the system's accuracy.  This result shows that calibration does positively correlate with accuracy.

\newpage
\subsection{Class-wise performance of the simulated MLPMixer expert for HAM10000}\label{ref:ham10000_mlpmixer_expert_perf}
 \begin{table}[h]
\centering
  \begin{tabular}{l | c c c c c c c | c}
    \multicolumn{7}{c}{\;\;\;\;\;\;\;\;Classes} \\
      metric & bkl & df & mel & nv & vasc & akiec & bcc  & weighted avg\\
      \hline
      precision & 0.52 & 0.33 & 0.51 & 0.82 & 0.27 & 0.44 & 0.47 & 0.71\\
      recall & 0.37 & 0.06 & 0.21 & 0.95 & 0.48 & 0.39 & 0.45 & 0.74\\
      f1-score & 0.43 & 0.10 & 0.30 & 0.88 & 0.34 & 0.41 & 0.46 & 0.71 \\
      
  \end{tabular}
  \caption{\textit{Performance of simulate MLPMixer Expert on HAM10000.} We can see that the trained model has non-uniform performance across different classes. The resulting model is still a valid simulation of real world expert who might be expert for some classes(class nv for example).}
  \label{tab:perf_mlp_mixer}
  \end{table}

\section{Additional Information about the Methods}\label{sec:baselines}
In this section, we provide additional implementation details for our comparison systems. We first note that the differentiable-triage algorithm \cite{okati2021differentiable} considers the triage level(or \textit{budget}) in the training of the algorithm. None of the other baselines have this aspect. Thus, to fairly compare all the other methods with the differentiable-triage algorithm, we use the same methodology employed by \citet{okati2021differentiable} in their paper (We refer the reader to Appendix C of their paper for more details). For each of the method, we also provide the details below:
\begin{enumerate}
    \item Softmax Surrogate \cite{pmlr-v119-mozannar20b}: for a \textit{budget} $b$ and the samples size $\mathcal{D}$, it sorts the samples in increasing order of $\max_{k \in [K]}p_{k}(\vx) - p_{\bot}(\vx)$, and then defers the $\min\left(\lfloor b\vert\mathcal{D}\vert\rfloor, n_{c}\right)$ where $n_{c}$ is the number of samples for which $p_{\bot}(\vx) \geq \max_{k \in [K]}p_{k}(\vx)$.
    
    \item One-Vs-All Surrogate: we use the same procedure as the softmax surrogate. 
    
    \item Score Baseline \cite{raghu2019algorithmic}: this method first trains a classifier model, and uses the classifier's predictive uncertainty to defer to the expert. Note that this classifier is trained in a regular way, i.e. it doesn't employ any additional procedure for deferral. During test time, it first sorts the dataset of size $\vert\mathcal{D}\vert$ in the increasing order of $\max_{k\in [K]}p_{k}(\vx)$, and defers to the expert first $\lfloor b\vert\mathcal{D}\vert\rfloor$ for the \textit{budget} $b$. The performance of this method depends on the reliability of the uncertainty estimates the classifier provides. We, therefore, use a post-processing calibration technique called Temperature Scaling \cite{10.5555/3305381.3305518} to calibrate the classifier using the validation dataset split. 
    
    \item Confidence Baseline \cite{Bansal2021IsTM}: this method first estimates $p(y=m)$, the probability of the expert being correct. However, this estimate is independent of the input sample $\vx$, i.e. $p(y=m|\vx) = p(y=m)$. Having obtained this estimate, it trains the system sequentially where at each iteration, it uses only $\min\left(\lfloor b\mathcal{D} \rfloor, n_{c}\right)$ samples with the lowest value of $p(y=m) - \max_{k \in [K]}p_{k}\left(\vx\right)$ in the corresponding mini-batch for training. Here, $n_{c}$ is the number of samples where $p(y=m) > \max_{k \in [K]}p_{k}\left(\vx\right)$. During test time for the \textit{budget} $b$, it first sorts the dataset of size $\vert\mathcal{D}\vert$ in the increasing order of $\max_{k \in [K]}p_{k}\left(\vx\right)$, and defer the first $\min\left(\lfloor b \vert\mathcal{D}\vert \rfloor, n_{c}\right)$ samples to the expert, where $n_{c}$ denotes the same quantity as before except this time for the test set samples.
    
    \item Differentiable Triage \cite{okati2021differentiable}: this is a sequential learning algorithm that first estimates the predictive model for a given \textit{budget} $b$, and then having learned the model, it approximates the optimal triage policy for the learned model and $b$. The optimal triage policy is to compare the model's prediction loss and the expert's prediction loss, and defer to the expert if the latter is smaller than the former. Therefore, the training algorithm assumes access to the expert's predictive loss as opposed to just the expert's predictions for the surrogate loss methods. Following the original authors, we use the Negative Log-Likelihood loss as the expert's loss. At test time, it use the learned approximation of the optimal triage policy to defer to the expert. 
    \end{enumerate}

\section{Additional Experimental Details}\label{sec:exp_details}
Below we provide more details on our experimental set-up. 

\paragraph{\texttt{CIFAR-10}} For the experiments on \texttt{CIFAR-10}, we use 28-layer Wide Residual Networks \cite{zagoruyko2016wide} without using any data augmentation techniques following \citet{pmlr-v119-mozannar20b}. We use SGD with a momentum of $0.9$, weight decay $5e-4$, and initial learning rate of $0.1$. We further use cosine annealing learning rate schedule. We monitor validation loss, and employ early stopping to terminate the training if the loss doesn't improve for $20$ epochs. The datasets are standardized to have $0$ mean and unit variance. We train the models with a batch size of 1024. These experimental settings apply to both the Softmax Surrogate and the One-vs-All surrogate loss. 

\paragraph{\texttt{HAM10000}} To simulate the expert, we train an 8-layer MLPMixer model \cite{Tolstikhin2021MLPMixerAA}. We make use of the publicly available code \footnote{\texttt{https://github.com/jaketae/mlp-mixer/}} for MLPMixer model. We resize the \texttt{HAM10000} images to $224\times224$ for our experiments. The 8-layer model has patch size of 16, expansion factor 2, and the dimensionality of the features to be 128. We train this model with Adam optimization algorithm with a learning rate of $0.001$, weight decay of $5e-4$. We further use cosine annealing learning rate schedule with a warm-up period of $5$ epochs. The model is trained with a batch size of $1024$, again with early stopping with a patience of $20$ epochs. Since our goal was to simulate the real-world expert, we did not do extensive hyperparameter search for the expert model. For our main model on \texttt{HAM10000}, we finetune ResNet34 model. The training settings are same for the surrogate loss methods for \texttt{CIFAR-10} experiments. 

For our other baselines, we use the code made available by the respective authors.

\end{document}

%% file: math_commands.tex
\usepackage{amsmath,amsfonts,bm}
\usepackage{upgreek}

\def\eqref#1{equation~\ref{#1}}

\def\1{\bm{1}}

\def\rgamma{{\upgamma}}
\def\rphi{{\upphi}}
\def\rpsi{{\uppsi}}

\def\rsm{{\textnormal{m}}}

\def\ry{{\textnormal{y}}}

\def\rvx{{\mathbf{x}}}

\def\vx{{\bm{x}}}

\DeclareMathAlphabet{\mathsfit}{\encodingdefault}{\sfdefault}{m}{sl}
\SetMathAlphabet{\mathsfit}{bold}{\encodingdefault}{\sfdefault}{bx}{n}

\DeclareMathOperator*{\argmax}{arg\,max}
\DeclareMathOperator*{\argmin}{arg\,min}

\DeclareMathOperator{\sign}{sign}